\documentclass[twoside,11pt]{article}

\usepackage{jmlr2e}
\usepackage{hyperref}

\usepackage{amsmath}             

\usepackage{algorithm}
\usepackage{algpseudocode}

\usepackage{xcolor}

\usepackage{graphicx}
\usepackage{natbib}
\usepackage{url} % not crucial - just used below for the URL 
\usepackage[normalem]{ulem} % for strikeout

\newcommand{\pkg}[1]{{\fontseries{b}\selectfont #1}} 

\newcommand{\alref}[1]{Algorithm~\ref{#1}}
\newcommand{\aref}[1]{Appendix~\ref{#1}}
\newcommand{\eref}[1]{Eq.~(\ref{#1})}
\newcommand{\fref}[1]{Figure~\ref{#1}}
\newcommand{\frefs}[1]{Figures~\ref{#1}}
\newcommand{\sref}[1]{Section~\ref{#1}}

\newcommand{\tref}[1]{Table~\ref{#1}}

\newcommand{\setVar}{\boldsymbol{X}}
\newcommand{\setVarexample}{\boldsymbol{Z}}
\newcommand{\setVarexamplet}{\boldsymbol{W}}
\newcommand{\bnet}{\mathcal{B}}
\newcommand{\graph}{\mathcal{G}}
\newcommand{\searchdg}{\mathcal{H}}
\newcommand{\searchdgadj}{H}
\newcommand{\searchdgpa}{\boldsymbol{h}}
\newcommand{\searchdgpal}{h}
\newcommand{\party}{\lambda}
\newcommand{\Party}{\Lambda}
\newcommand{\permy}{\pi}

\newcommand{\chainStep}{t}
\newcommand{\pars}{\theta}
\newcommand{\Pa}{\mbox{{\bf Pa}}}
\newcommand{\obs}{D}
\newcommand{\score}{S}
\newcommand{\sumscore}{\Sigma}
\newcommand{\sumscoret}{\tilde{\Sigma}}
\newcommand{\tf}{\bar{f}}
\newcommand{\tg}{\tilde{g}}
\newcommand{\maxscore}{M}
\newcommand{\order}{\prec}
\newcommand{\neibr}[1]{\#\mathrm{nbd}(#1)}

\newcommand{\neibi}[1]{\mathrm{nbd}^{i}(#1)}
\newcommand{\neibj}[1]{\mathrm{nbd}^{j}(#1)}
\newcommand{\Sord}{R}
\newcommand{\Sordmax}{Q}
\newcommand{\Nobs}{N}
\newcommand{\Nbold}{\boldsymbol{N}}

\newcommand{\BiDAG}{\pkg{BiDAG} }

\ShortHeadings{Efficient Sampling and Structure Learning of Bayesian Networks}{Kuipers, Suter and Moffa}
\firstpageno{1}

\begin{document}

\title{Efficient Sampling and Structure Learning \\ of Bayesian Networks\thanks{\pkg{R} package \pkg{BiDAG} is available at \url{https://CRAN.R-project.org/package=BiDAG}}}

\author{\name Jack Kuipers \email jack.kuipers@bsse.ethz.ch
\\[0.5ex] \name Polina Suter
\\ \addr D-BSSE, ETH Zurich, Mattenstrasse 26, 4058 Basel, Switzerland
\AND
\name Giusi Moffa \\ \addr Division of Psychiatry, University College London, London, UK \\
Department of Mathematics and Computer Science, University of Basel, Basel, Switzerland}

\editor{\hspace{-1.5cm}{\color{white}\rule[-0.2cm]{2cm}{0.6cm}} \vspace{-1.8cm}}

\maketitle

\begin{abstract}%   <- trailing '%' for backward compatibility of .sty file
Bayesian networks are probabilistic graphical models widely employed to understand dependencies in high dimensional data, and even to facilitate causal discovery.  Learning the underlying network structure, which is encoded as a directed acyclic graph (DAG) is highly challenging mainly due to the vast number of possible networks in combination with the acyclicity constraint.  Efforts have focussed on two fronts: constraint-based methods that perform conditional independence tests to exclude edges and score and search approaches which explore the DAG space with greedy or MCMC schemes.  Here we synthesise these two fields in a novel hybrid method which reduces the complexity of MCMC approaches to that of a constraint-based method.  Individual steps in the MCMC scheme only require simple table lookups so that very long chains can be efficiently obtained.  Furthermore, the scheme includes an iterative procedure to correct for errors from the conditional independence tests.  The algorithm offers markedly superior performance to alternatives, particularly because DAGs can also be sampled from the posterior distribution, enabling full Bayesian model averaging for much larger Bayesian networks.
\end{abstract}

     
\section{Introduction}\label{intro}

Bayesian networks are statistical models to describe and visualise in a compact graphical form the probabilistic relationships between variables of interest. The nodes of a graphical structure correspond to the variables, while directed edges between the nodes encode conditional independence relationships between them. The most important property of the digraphs underlying a Bayesian network is that they are acyclic, i.e.\ there are no directed paths which revisit any node.  Such graphical objects are commonly denoted DAGs (directed acyclic graphs).

Alongside their more canonical use for representing multivariate probability distributions, DAGs also play a prominent role in describing causal models \citep{greenland99causal, pearl00, hr06instruments, wr07four} and facilitating causal discovery from observational data \citep{mkb09, moffaetal17}, though caution is warranted in their use and interpretation \citep{dawid10}.  However, the potential for uncovering the mechanisms underlying scientific phenomena in disciplines ranging from the social sciences \citep{elwert13graphical} to biology \citep{friedmanetal00, friedman04, kuipersetal18} has driven interest in DAG inference.

To fully characterise a Bayesian network we need the DAG structure and the parameters of an associated statistical model which explicitly describes the probabilistic relationships between the variables. Given sample data from the joint probability distribution on the node variables, learning the graphical structure is in general the more challenging task.  The difficulty mostly rests with respecting acyclicity in combination with the sheer size of the search space, which grows super-exponentially with the number of nodes $n$, since the logarithm grows quadratically \citep{robinson70, robinson73}. A curious illustration of this growth is that the number of DAGs with 21 nodes approximately equals the estimated number of atoms in the observable universe ($\approx10^{80}$).

\subsection{Bayesian network notation}

Bayesian networks represent a factorisation of multivariate probability distributions of $n$ random variables $\setVar = \{ X_1, \ldots, X_n\}$ by encoding conditional dependencies in a graphical structure.  A Bayesian network $\bnet = (\graph, \pars)$ consists of a DAG $\graph$ and a set of parameters $\pars$ which define the conditional distribution $P(X_i \mid \Pa_i)$ of each variable $X_i$ on its parents $\Pa_i$ in the graph. The distribution a Bayesian network represents is assumed to satisfy the   
\emph{Markov property} \citep[described for example in][]{bk:kf09} that each variable $X_i$ is independent of its non-descendants given its parents $\Pa_i$, allowing the joint probability distribution to factorise as
\begin{equation} \label{distfactorisation}
P( X_1, \ldots, X_n ) = \prod_i^n P(X_i \mid \Pa_i) \,
\end{equation}
Learning the parameters $\pars$ which best describe a set of data $\obs$ for a given graph $\graph$ is generally straightforward for complete data, while learning the structural dependence in $\setVar$ and the DAG $\graph$ itself constitutes the main difficulty. 

Due to the symmetry of conditional independence relationships, the same distribution might factorise according to different DAGs. DAGs encoding the same probability distribution constitute an equivalence class: they share the v-structures \citep[two unconnected parents of any node,][]{vp90} and the skeleton (the edges if directions were removed).  The equivalence class of DAGs can be represented as an essential graph \citep{amp97} or a completed partially directed acyclic graph (CPDAG) \citep{chickering02CPDAG}.   Based purely on probabilistic properties, Bayesian networks can therefore only be learned from data up to an equivalence class. 

\subsection{DAG posteriors} \label{sec:DAGposterior}

A further challenge to learning the graph structure (or its equivalence class) is accounting for the uncertainty in the structure and parameters given the data. A natural strategy in Bayesian inference consists of sampling and averaging over a set of similarly well fitting networks.  Each DAG $\graph$ receives a score equal to its posterior probability given the data $\obs$
\begin{equation} \nonumber
P( \graph \mid \obs) \propto P(\obs \mid \graph) P(\graph) \,
\end{equation}
where the likelihood $P(\obs \mid \graph)$ has been marginalised over the parameter space.  When the graph and parameter priors satisfy certain conditions of structure modularity, parameter independence and parameter modularity \citep{hg95,friedmanetal00,fk03} the score decomposes as
\begin{equation} \label{scoredecomp}
P( \graph \mid \obs) \propto P(D \mid \graph) P(\graph) = \prod_{i=1}^{n} \score(X_i, \Pa_i \mid \obs) \, ,
\end{equation}
involving a function $\score$ which depends only on a node and its parents.  For discrete categorical data, a Dirichlet prior is the only choice satisfying the required conditions for decomposition, leading to the BDe score of \cite{hg95}. For continuous multivariate Gaussian data, the inverse Wishart prior leads to the BGe score (\citealp{gh02}; corrected in \citealp{kmh14}).  In this manuscript we focus on the continuous case with the BGe score when evaluating the complexity of our approach, and discuss the discrete categorical case in \aref{catdata}.

\subsection{State of the art structure learning}

Traditional approaches to structure learning fall into two categories (and their combination): 
\begin{itemize}
\item constraint-based methods, relying on conditional independence tests
\item score and search algorithms, relying on a scoring function and a search procedure 
\end{itemize}

For each class we provide below a brief review of concepts and algorithms pertinent to this work. 

\subsubsection{Constraint-based methods}

Constraint-based methods exploit the property of Bayesian networks that edges encode conditional dependencies. If data show that a pair of variables are independent of each other when conditioning on at least one set (including the empty set) of the remaining variables, then we can exclude a direct edge between the corresponding nodes in the graph.  The most prominent example of constraint-based methods is the well known PC algorithm of \cite{bk:sgs00}, more recently popularised by \cite{art:KalischB2007}, who provided consistency results for the case of sparse DAGs and an \pkg{R} \citep{R17} implementation within the \pkg{pcalg} package \citep{art:KalischMCMB2012}. 

Rather than exhaustively searching the $2^{(n-2)}$ possible conditioning sets for each pair of nodes, the crucial insight of the PC algorithm is to perform the tests in order of increasing complexity.  Starting from a fully connected (undirected) graph, the procedure tests marginal independence (conditioning on the empty set) for all pairs of nodes.  Then it performs pairwise conditional independence tests between pairs of nodes which are still directly connected, conditioning on each adjacent node of either node in the pair, and so on conditioning on larger sets.  Edges are always deleted when a conditional independence test cannot be rejected.  This strategy differs from the typical use of hypothesis testing since it assumes edges to be present by default, but with conditional independence as the null hypothesis. 

Edges which are never deleted through the process form the final skeleton of the PC algorithm. The conditional independencies which are not rejected identify all v-structures of the graph, fixing the direction of the corresponding edges. At this point it may still be possible to orient some edges, to ensure that no cycles are introduced and no additional v-structures are created \citep{meek95}. The algorithm finally returns the CPDAG of the Markov equivalence class consistent with the conditional dependencies compatible with the data.

In the final skeleton, for each node the remainder of its adjacent neighbourhood will have been conditioned upon. For the node with largest degree $K$, at least $K2^{K-1}$ tests will have been performed and the algorithm is of exponential complexity in $K$.  In the best case it may still run with $K$ around 25--30, but in the worst case the base can increase giving a complexity bound of $O(n^K)$ making the algorithm infeasible even for low $K$ for large DAGs \citep{art:KalischB2007}. 

For sparse large graphs, the PC algorithm can be very efficient.  Despite its efficiency, because it can only delete edges by performing a large number of (correlated) independence tests, the PC algorithm tends to have a high rate of false negatives and only find a fraction of the edges in the true network \citep{uhleretal13}.  Increasing the threshold for the conditional independence tests does little to alleviate the problem of false negatives while increasing runtime substantially.  Another consequence of repeated tests is that the output of the PC algorithm can depend on the order of the tests (or the ordering of the input data), leading to unstable estimates in high dimensions, and modifications have been advanced to mollify this effect \citep{cm14}.

\subsubsection{Score and search methods}

On the other side of the coin are score and search methods and MCMC samplers. Each DAG gets a score, typically a penalised likelihood or a posterior probability (\sref{sec:DAGposterior}).  An algorithm then searches through the DAG space for the structure which optimises the score. Further, algorithms which can return a collection of DAGs, sampled proportionally to their score, enable a more complete characterization of the uncertainty surrounding structure learning. 

The most basic sampler is structure MCMC where each step involves adding, deleting or reversing an edge \citep{my95,gc03} and accepting the move according to a Metropolis-Hastings probability.  The scheme is highly flexible -- amplifying the score leads to simulated annealing while modulating the amplification through the acceptance rate leads to adaptive tempering, to speed up the traversal and exploration of the DAG space.  One way to speed up convergence is to sample from the neighbourhood of DAGs with one edge added, removed or reversed, in proportion to the structure score \citep[as for example in][]{jc18}. For optimisation only, greedy hill climbing chooses the highest scoring DAG in the neighbourhood as the new starting point.  A popular algorithm for greedy search on the space of Markov equivalent DAGs is greedy equivalence search (GES) \citep{chickering02}.

When the score is decomposable (as in \eref{scoredecomp}), we only need to rescore nodes whose parents change and structure based methods enjoy an $O(n)$ speedup.   Since the decomposability in \eref{scoredecomp} mimics the factorisation in \eref{distfactorisation} commonly used scores like the BIC penalised likelihood or the BDe or BGe possess this property, which is essential for more advanced algorithms.

Order MCMC \citep{fk03} spearheaded a path to reduce the search space by combining large collections of DAGs together, namely all DAGs sharing the same topological ordering of the nodes. The score of the order consists of the sum of the scores of all DAGs consistent with it, and an MCMC scheme runs on the space of orders which are simply permutations of the $n$ nodes.  Although the number of DAGs compatible with each order also grows super-exponentially, the sum of all their scores involves $\approx 2^{n}$ different contributions and can be evaluated in exponential time \citep{buntine91}. For larger $n$ computations become quickly prohibitive and imposing a hard limit $K$ on the number of parents allowed for each node is common to artificially reduce the complexity to polynomial $O(n^{K+1})$.  Even so, the strategy of combining large sets of DAGs to work on the much smaller (though still factorial) space of orders, enormously improves convergence upon structure MCMC, enabling the search and sampling of much larger graphs (for moderate or low $K$).

Score decomposability is also necessary for order-based greedy search \citep{tk05} as well as for dynamic or integer linear programming approaches \citep{ks04,em07,htw16,cussens11,cussensetal17} to structure learning. One limitation of order MCMC for Bayesian model averaging, is that DAGs may belong to multiple orders, introducing a bias in the sampling. To avoid the bias we can work on the space of ordered partitions \citep{km17} which provide a unique representation of each DAG.  Other MCMC alternatives include large scale edge reversal \citep{gh08} and Gibbs sampling \citep{gm16} moves.  Unbiased MCMC schemes, such as these, are currently the only viable approaches to sampling and accounting for structure uncertainty, though still limited to smaller or relatively sparse graphs.  

As the size and connectivity of the target DAGs increase, the wide spectrum of constraint-based and score and search algorithms, cannot but fail to converge or discover optimally scoring graphs.  To limit the extent to which the search space grows with the number of nodes, \cite{fnp99} pruned it by only allowing edges from selected candidate parents and performing a greedy search in the restricted search space.  They also iteratively updated the set of candidate parents based on the current best DAG discovered. With the same aim of limiting the search space, \cite{tba06} brought together the ease of conditional independence testing and the performance of DAG searching, to benefit from their individual advantages.   First a constraint-based method, akin to the PC algorithm, identifies a (liberal) undirected skeleton.  A greedy search then acts on the restricted search space defined by excluding edges not included in the reference skeleton. Since score and search, when feasible, tends to perform better \citep{heckerman2006bayesian} than constraint-based methods, the hybrid approach of \cite{tba06} outperformed previous methods.  \cite{nhm18} recently investigated the consistency properties of hybrid approaches using GES in high dimensional settings.

\subsection{Original contribution}

In this work, we bring the power and sophistication of order and partition MCMC to the hybrid framework for structure learning. The result is a highly efficient algorithm for search and sampling with marked improvements on current state of the art methods.  The key is to observe that the exponential complexity in $K$ of $n^{K}$ for order or partition MCMC \citep{fk03,km17} derives from allowing among the potentially $K$ parents of each node any of the other $(n-1)$. If the set of $K$ potential parents is pre-selected, for example through a constraint-based method relying on conditional independence tests, the space and time complexity reduces to $2^K$ for the search of an optimal structure, and $3^K$ for unbiased structure sampling.   The complexity of the search then matches that of the testing component of the PC algorithm.  Along with the standard pre-computation of parents set scores, which are exponentially costly, we introduce a method to also precompute tables of partial sums of parent set scores. For order-based sampling, there is no complexity overhead, and the storage space is the same as the standard score tables.  In particular we tabulate every score quantity needed for the MCMC scheme.  During each MCMC step we then simply need to look up the relevant scores providing a very efficient sampler.

A distinctive feature of our method is not to fully trust the search space provided by the initial constraint-based method. Each node is entitled to have as parent any of the permissible nodes in the search space, and an additional arbitrary one from outside that set. Accounting for the expansion to the potential parent set, each MCMC step takes an expected time of order $O(K)$, despite scoring vast sets of DAGs at a time. Therefore the complexity is comparable or even lower than that of structure MCMC moves \citep[between $O(n)$ and $O(n^2)$, though some rejected moves can be $O(1)$;][]{gc03}.  The expansion beyond the skeleton provides a mechanism to iteratively improve the search space until it includes the maximally scoring DAG encountered, or the bulk of the posterior weight.  Extensive simulations show that our method strongly outperforms currently available alternatives, enabling efficient inference and sampling of much larger DAGs. 

In \sref{sec:order} we develop efficient algorithms for order-based sampling for a known search space and examine their convergence.  In \sref{sec:spacemax} we demonstrate how to expand the search space iteratively, and show how it notably improves performance in finding the best DAG.  Finally, we extend the scheme to the space of partitions to provide an unbiased sampler in \sref{sec:partition}, with discussions in \sref{discussion}.  All algorithms introduced here are implemented in the \pkg{R} \citep{R17} package \href{https://CRAN.R-project.org/package=BiDAG}{\BiDAG}.

\section{Order-based DAG sampling on a given search space} \label{sec:order}

\begin{figure}[t]
  \centering
  \includegraphics[width=\textwidth]{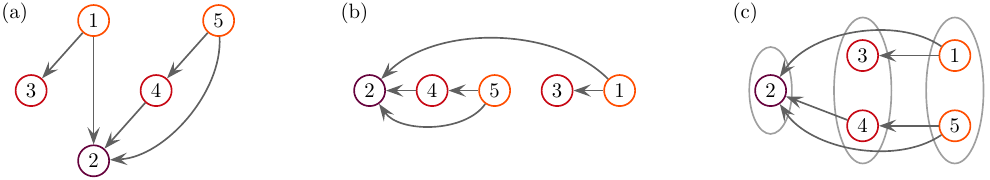}
  \caption{The DAG in (a) is compatible with the order depicted in (b), as edges only originate from nodes further right in the chain. The same DAG is also compatible with 8 other orders.  The DAG however can be uniquely assigned to a labelled partition by collecting outpoints into blocks we will call blocks to arrive at the representation in (c).}
  \label{fig:dagsorderpartition}
\end{figure}

The order MCMC algorithm of \cite{fk03} requires arranging the $n$ nodes of a DAG in topological order $\order$.  We associate a permutation $\pi_{\order}$ with each order.  For a DAG to be compatible with an order, the parents of each node must have a higher index in the permutation
\begin{equation}
\graph \in \order  \overset{\mathrm{def}}{\iff} \forall i \, , \; \forall \left\{j : X_j \in \Pa_i \right\} , \;   \pi_{\order}[i] < \pi_{\order}[j]
\end{equation}
Visually, when we place the nodes in a linear chain from left to right according to $\pi_{\order}$, edges may only come from nodes further to the right (\fref{fig:dagsorderpartition}b).  With the rows and columns labelled following $\pi_{\order}$, the adjacency matrix of a compatible DAG is lower triangular so that a total of $2^{\binom{n}{2}}$ DAGs are compatible with each order.  

\subsection{Order MCMC}

The idea of order MCMC is to combine all DAGs consistent with an order and reduce the problem to the much smaller space of permutations instead of working directly on the space of DAGs.  Each order $\order$ receives a score $\Sord(\order \mid \obs)$ equal to the sum of the scores of all DAGs in the order
\begin{equation}
\Sord(\order \mid \obs) = \sum_{\graph \in \order} P( \graph \mid \obs) \,
\end{equation}
The strategy is to build a Markov chain on the space of orders \citep{fk03}.  From the order $\order_\chainStep$ at the current iteration $\chainStep$, a new order $\order'$ is proposed and accepted with probability
\begin{equation} \label{orderMHaccprob}
\rho =  \min \left \{ 1 , \frac{\Sord(\order' \mid \obs) }{\Sord(\order_\chainStep \mid \obs)} \right \} \,
\end{equation}
to provide a chain with a stationary distribution proportional to the score $\Sord(\order \mid \obs)$ for symmetric proposals. The standard proposal is a \emph{global swap} of two nodes in the order while leaving all others fixed. A more local move consists of transposing two adjacent nodes in a \emph{local transposition}.

Instead of naively scoring all DAGs in an order, \cite{fk03} used the factorisation in \eref{scoredecomp}
\begin{equation}\label{orderScore}
\Sord(\order \mid \obs) \propto \sum_{\graph \in \order} \prod_{i=1}^{n} \score(X_i, \Pa_i \mid \obs) = \prod_{i=1}^{n} \sum_{\Pa_i \in \order} \score(X_i, \Pa_i \mid \obs) \,
\end{equation}
to exchange the sum and product \citep[following][]{buntine91}.  The sum is restricted to parent subsets compatible with the node ordering
\begin{equation}
\Pa_i \in \order  \overset{\mathrm{def}}{\iff} \forall \left\{j : X_j \in \Pa_i \right\} , \;   \pi_{\order}[i] < \pi_{\order}[j]
\end{equation}

The score of the order therefore reduces to sums over all compatible parent subsets, eliminating the need of summing over DAGs.  For a node with $k$ possible parents further along the order, there are $2^k$ parents subsets.  Evaluating the score of the order therefore requires $\sum_{k=0}^{n-1}2^{k}=(2^{n}-1)$ evaluations of the score function $\score$.  This provides a massive reduction in complexity compared to scoring all $2^{\binom{n}{2}}$ DAGs in the order individually.  The exponential complexity in $n$ is still too high for larger DAGs so a hard limit $K$ on the size of the parent sets is typically introduced to obtain polynomial complexity of $O(n^{K+1})$ evaluations of $\score$.  For larger DAGs however, $K$ must be rather small in practice, so that the truncation runs the risk of artificially excluding highly scoring DAGs.  As a remedy, we start by defining order MCMC on a reduced search space, for example selected on the basis of prior subject knowledge or from a skeleton derived through constraint-based methods.

\subsection{Restricting the search space} 

The search space can be defined by a directed (not necessarily acyclic) graph $\searchdg$, or its adjacency matrix, $\searchdgadj$:
\begin{equation}
\searchdgadj_{ji} = 1 \mbox{ if } \{j,i\} \in \searchdg \,
\end{equation}
One advantage with respect to simply using an undirected skeleton, which corresponds to a symmetric matrix, is that the directed graph naturally allows for the inclusion of prior information about edge directionality.   Prior beliefs about undirected edges can also be included by treating both directions equally.  In the search space, each node has the following set of permissible parents
\begin{equation}
\searchdgpa^i = \{X_j : H_{ji} =1 \} \, 
\end{equation}
For a set of size $K$ we follow the standard practice to evaluate the score of each possible combination and store them in a table (as in the example on the left of \tref{scoretableexample}).  Since there are $2^{K}$ possible combinations, and for the BGe score each involves taking the determinant of a matrix, the complexity of building this table is $O(K^{3}2^{K})$.  For indexing we use a one-to-one mapping between parent subsets and integers; specifically the following binary mapping:
\begin{equation}
f(\setVarexample)=\sum_{j=1}^{K}I(\searchdgpal^{i}_j \in \setVarexample)2^{j-1}
\end{equation}
using the indicator function $I$.

\subsection{Efficient order scoring} 

To score all the DAGs compatible with a particular order we still need to select and sum the rows in the score tables where the parent subset respects the order
\begin{equation}\label{orderScoresdg}
\Sord_{\searchdg}(\order \mid \obs) \propto \prod_{i=1}^{n} \sum_{\substack{\Pa_i \subseteq \searchdgpa^i \cr \Pa_i \in \order} } \score(X_i, \Pa_i \mid \obs) \,
\end{equation}
with the additional constraint that all elements in the parent sets must belong to the search space defined by $\searchdg$.  From the precomputed score table of all permissible parent subsets in the search space, we select those compatible with the order constraint.  Simply running through the $2^K$ rows takes exponential time (in $K$) for each node.  Unlike other order-based schemes, we can avoid this by building a second table: the summed score table (see the example on the right of \tref{scoretableexample}).

In \aref{sumtableapp} we detail the algorithmic steps which allow us to build the summed score table for each variable with a complexity of $O(K^{2}2^{K})$, adding no overhead with respect to the complexity $O(K^{3}2^{K})$ of building the original score table. The size of the summed score tables is the same as that of the original score tables and hence require the same storage space, while providing the means to efficiently score each order.  For each node we look up the relevant row in the summed score table and by moving linearly along the order we can compute \eref{orderScoresdg} in $O(Kn)$ as
\begin{equation}\label{orderScoresdgtable}
\Sord_{\searchdg}(\order \mid \obs) \propto \prod_{i=1}^{n} \sumscore^{i}_{f(\substack{\searchdgpa^i \in \order})} \,
\end{equation}
where
\begin{equation}
\searchdgpa^i \in \order \; \overset{\mathrm{def}}{=} \; \left\{X_j \in \searchdgpa^i : \pi_{\order}[i] < \pi_{\order}[j] \right\}
\end{equation}
are the elements of $\searchdgpa^i$ compatible with the order constraints and 
\begin{equation}
\sumscore^{i}_{f(\substack{\searchdgpa^i \in \order})} = \sum_{\substack{\Pa_i \subseteq \searchdgpa^i \cr \Pa_i \in \order} } \score(X_i, \Pa_i \mid \obs)
\end{equation}
is the sum of scores of all the parent sets in the search space respecting the order, precomputed using \alref{parentsumalg}. With the restriction through $\searchdg$ we use $\Sord_{\searchdg}(\order \mid \obs)$ in the MCMC scheme (\eref{orderMHaccprob}).

\subsection{A Gibbs move in order space} \label{sec:Gibbsmove}

\begin{figure}[t]
  \centering
  \includegraphics[width=0.5\textwidth]{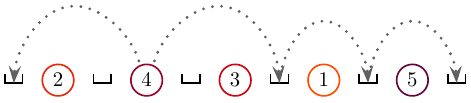}
  \caption{For any randomly chosen node, here 4, we score the entire neighbourhood of new positions in the order by performing a sequence of local transpositions and including the known score of the current position.  The new placement is sampled proportionally to the scores of the different orders in the neighbourhood.}
  \label{ordermovepic}
\end{figure}

On the space of orders we define a new \emph{node relocation} move: from the current state in the chain, $\order_\chainStep$, first sample a node uniformly from the $n$ available, say node $i$.  For the move sample a new position for node $i$ conditional on keeping the relative ordering of the remaining nodes unchanged.  Define the neighbourhood of the order under this move, $\neibi{\order_\chainStep}$, to be all orders with node $i$ placed elsewhere in the order or at its current position, as in the example in \fref{ordermovepic} with node 4 chosen. Finally sample a proposed order $\order'$ proportionally to the scores of all orders in the neighbourhood.  As a consequence the move is always accepted and the next step of the chain set to the proposed order $\order_{\chainStep+1}=\order'$.  We summarise the move in \alref{ordermovealg}.

\begin{algorithm}[t]
\caption{Node relocation move in the space of orders}\label{ordermovealg}
\begin{algorithmic}
\State \textbf{input} The order $\order_\chainStep$ at step $\chainStep$ of the chain
\State Sample node $i$ uniformly from the $n$
\State Build and score all orders in the neighbourhood $\neibi{\order_\chainStep}$, \\$\quad$ through consecutive local transpositions of node $i$
\State Sample proposed order $\order'$ from $\neibi{\order_\chainStep}$ proportionally to $\Sord_{\searchdg}(\order' \mid \obs)$
\State Set $\order_{\chainStep+1}=\order'$
\State \textbf{return} $\order_{\chainStep+1}$
\end{algorithmic}
\end{algorithm}

The newly defined node relocation move satisfies detailed balance
\begin{equation} \label{detailedbalance}
P(\order' \mid \order) \Sord_{\searchdg}(\order \mid \obs)  = P(\order \mid \order') \Sord_{\searchdg}(\order' \mid \obs)   \,
\end{equation}
where $P(\order' \mid \order)$ is the transition probability from $\order$ to $\order'$.  The transition involves first sampling a node $i$ and then the order proportionally to its score so that (for orders not connected by a local transposition)
\begin{equation} \label{transprobeqn}
P(\order' \mid \order) = \frac{1}{n}\frac{\Sord_{\searchdg}(\order' \mid \obs)}{\sum_{\order'' \in \neibi{\order}}\Sord_{\searchdg}(\order'' \mid \obs)} \,
\end{equation}
The reverse move needs the same node $i$ to be selected, and as $\neibi{\order}=\neibi{\order'}$ the denominators cancel when substituting into \eref{detailedbalance}.  For orders connected by a local transposition, say node $i$ swapped with the adjacent node $j$, there are two possible paths connecting the orders and a transition probability of
\begin{equation} \label{transprobadjeqn}
P(\order' \mid \order) = \frac{1}{n}\frac{\Sord_{\searchdg}(\order' \mid \obs)}{\sum_{\order'' \in \neibi{\order}}\Sord_{\searchdg}(\order'' \mid \obs)} + \frac{1}{n}\frac{\Sord_{\searchdg}(\order' \mid \obs)}{\sum_{\order'' \in \neibj{\order}}\Sord_{\searchdg}(\order'' \mid \obs)}\,
\end{equation}
Since the reverse move involves the same pair of nodes, we can again directly verify detailed balance by substituting into \eref{detailedbalance}.

The move is aperiodic since the original order is included in the neighbourhood. It is possible to reach any order from any other by performing $(n-1)$ steps making the chain also irreducible. Therefore the newly defined move satisfies the requirements for the chain to converge and provide order samples from a probability distribution proportional to the score $\Sord_{\searchdg}(\order \mid \obs)$.  

The node relocation move naturally provides a fixed scan Gibbs sampler by cycling through the nodes sequentially, rather than sampling at each step.

\subsection{Chain complexity}

In the previous sections we considered three types of move: \emph{global swap}, \emph{local transposition} and  \emph{node relocation}. In the global swap we need to rescore all nodes between the two swapped ones, since their set of permissible parents may change. The complexity of scoring the proposed order is $O(n)$. Moving from one order to any other is possible in $(n-1)$ steps (assuming non-zero order scores), making the chain irreducible.  In the case of local transposition we can rescore the proposed order in $O(1)$ and it takes $O(n^2)$ steps to gain access to any order.

The node relocation move has the same complexity of the standard global swap. To move through the full neighbourhood of size $n$, we can sequentially transpose node $i$ with adjacent nodes.  Since each local transposition takes a time $O(1)$ to compute the score of the next order, scoring the whole neighbourhood takes $O(n)$. 

For sampling we mix the three moves into a single scheme.  Since the global swap involves rescoring $\approx\frac{n}{3}$ nodes on average at each step, while the local transposition involves $2$ and the node relocation $2n$ we can keep the average complexity at $O(1)$ if we sample the more expensive moves with a probability $\propto \frac{1}{n}$.  In this way, we can also balance their computational costs. For simplicity we assign each move equal average computational time by selecting them with a probability of $(\frac{6}{n+7}, \frac{n}{n+7}, \frac{1}{n+7})$ respectively.  With the move mixture, the number of steps to reach any order is $O(n)$ so following the heuristic reasoning of \cite{km17} we would expect convergence of the chain to take $O(n^2\log(n))$ steps, a complexity consistent with our simulation results (\fref{fig:R2convergence}).

Once the score tables have been computed, the complexity of running the whole chain is also $O(n^2\log(n))$.  Utilising the lookup table of summed scores reduces the complexity substantially by a factor of $O(n^K)$ compared to standard order MCMC where one simply restricts the size of parent sets to $K$ and only utilises the basic score tables.

\subsection{DAG sampling}\label{sec:dagSampling}

Order MCMC produces a chain on the space of orders and one needs extra steps to build a sample of DAGs. To draw a DAG consistent with a given sampled order, we can sample the parents of each node proportionally to the entries respecting the order in the score table.  The complexity of drawing a DAG from an order remains exponential, of $O(2^{K})$, and thinning the scheme in a way that DAG sampling only happens periodically in the order chain may be appropriate. A convenient frequency should be such that the computational time for DAG sampling is at most comparable to running the order chain inbetween.  

\subsection{Extending the search space} \label{sec:expand}

A restricted search space, derived for example through constraint-based methods, may exclude relevant edges.  To address this problem, we extend our approach by softening the restrictions.  In particular we allow each node to have one additional parent from among the nodes outside its permissible parent set.  The score of each order becomes
\begin{equation}\label{orderScoresdgx}
\Sord_{\searchdg}^{+}(\order \mid \obs) \propto \prod_{i=1}^{n} \sum_{\substack{\Pa_i \subseteq \searchdgpa^i \cr \Pa_i \in \order} } \Bigg[\score(X_i, \Pa_i \mid \obs) + \sum_{\substack{X_j \notin \searchdgpa^i \cr \pi_{\order}[i] < \pi_{\order}[j]} }\score(X_i, \{\Pa_i, X_j\} \mid \obs) \Bigg] \,
\end{equation}
For the efficient computation of the sum, we build a score table for each node and each additional parent.  For a node with $K$ parents this leads to $(n-K+1)$ tables in total and we perform \alref{parentsumalg} of \aref{sumtableapp} on each of them.  The time and space complexity of building these tables is therefore an order $n$ more expensive than using the restricted search space.  Given the tables, however, a simple lookup is sufficient to score an order
\begin{equation}\label{orderScoresdgxtable}
\Sord_{\searchdg}^{+}(\order \mid \obs) \propto \prod_{i=1}^{n} \Bigg[ \sumscore^{i}_{f(\substack{\searchdgpa^i \in \order})} + \sum_{\substack{X_j \notin \searchdgpa^i \cr \pi_{\order}[i] < \pi_{\order}[j]} }\sumscore^{ij}_{f(\substack{\searchdgpa^i \in \order})} \Bigg] \,
\end{equation}
where we also index the summed scores with the additional parent
\begin{equation}
\sumscore^{ij}_{f(\substack{\searchdgpa^i \in \order})} = \sum_{\substack{\Pa_i \subseteq \searchdgpa^i \cr \Pa_i \in \order} } \score(X_i, \{\Pa_i, X_j\} \mid \obs) \, , \qquad X_j \notin \searchdgpa^i
\end{equation}
The complexity of scoring the order is $O(n^2)$. 

\subsubsection{Move complexity}

For the local transposition where we swap two adjacent nodes in the order, if neither is in the permissible parent set of the other we simply update one element of the sum in \eref{orderScoresdgxtable} in $O(1)$.  If either is in the permissible parent set, all terms need to be replaced in $O(n)$.  However since the nodes have up to $K$ parents, the probability of a permissible parent being affected is $\propto \frac{K}{n}$ giving an average complexity of $O(K)$.  For the global swap the maximum complexity is $O(n^2)$ when many of the intermediate nodes have among their permissible parents either of the swapped nodes, but on average the complexity is $O(Kn)$ following the same argument as above.  The node relocation move has the same complexity, so that the weighted mixture of moves typically takes just $O(K)$.

\begin{figure}[t]
  \centering
  \includegraphics[width=0.85\textwidth]{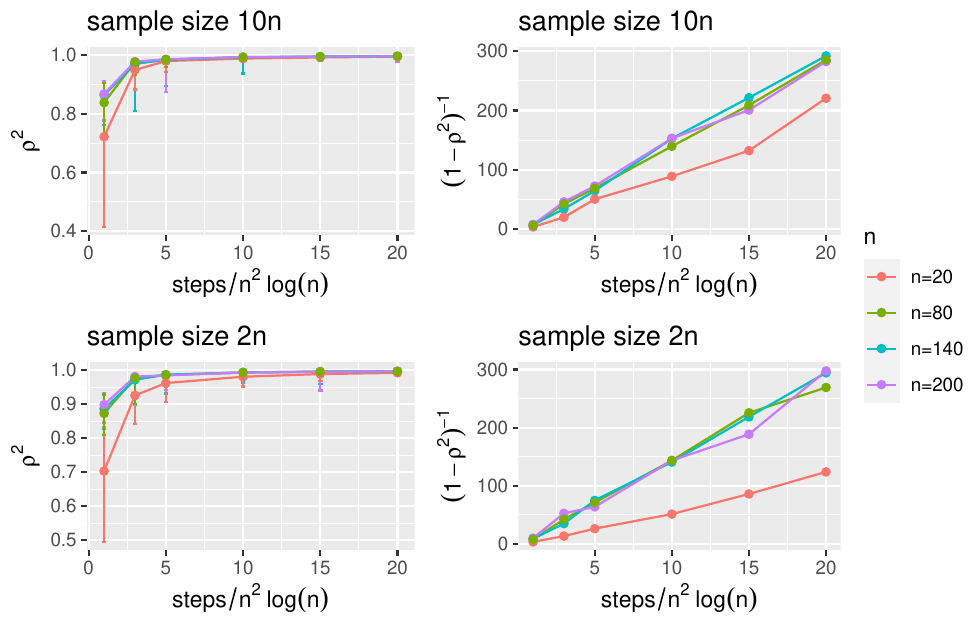}
  \caption{The correlation between edge probabilities from different runs as the size of the network increases.  The simulation setting is described in \aref{app:simulations}.  The number of steps in the chain is scaled by $n^2\log(n)$.  The transformation on the right highlights the roughly linear improvement in the convergence measure with the number of steps and that there is little dependence on the network size.}
  \label{fig:R2convergence}
\end{figure}

\subsection{Convergence} \label{sec:convergence}

To examine the convergence of our MCMC scheme, we ran simulations as described in \aref{app:simulations}, and for each simulation repetition we compared two independent runs with different random initial points in the final search space for each dataset. For the two runs we computed the squared correlation $\rho^2$ between the posterior probabilities of each edge, after excluding a burn-in period of 20\%.  Since most edges are absent, only edges with a posterior probability greater than 5\% in at least one run were included.  The median $\rho^2$ along with the first and third quartiles are displayed in \fref{fig:R2convergence}.  By scaling the number of MCMC steps with $n^2\log(n)$ we observe the correlation approaching 1.  To examine the behaviour in more detail, we can consider $(1-\rho^2)^{-1}$ which increases roughly linearly with the scaled number of steps.  We see, especially, that there is little dependence on the size of the network, apart from a slower convergence for the smallest network size.  With reasonable consistency, and importantly no obvious decrease in scaled performance as the number of variables $n$ increases, the simulation results of \fref{fig:R2convergence} are in line with the estimate of requiring $O(n^2\log(n))$ steps for the MCMC convergence.  Similar insensitivity to the network size is also visible for the root mean square difference between runs (\fref{fig:RMSDconvergence}).

\section{Search space and maximal DAG discovery}\label{sec:spacemax}

In order to sample DAGs effectively, our search space needs to cover the bulk of the posterior weight.  We describe here an iterative scheme to search for the highest scoring DAG in the current extended search space, and use it to update and improve the search space itself.

\subsection{Maximal DAG discovery}

In addition to sampling DAGs, we can also employ the MCMC scheme to search for the maximally scoring or maximum a posteriori (MAP) DAG.  To this end \citep[and analogously to][]{tk05} we replace the score of each order by the score of the highest scoring DAG in that order
\begin{equation}\label{orderScoresdgmax}
\Sordmax_{\searchdg}(\order \mid \obs) = \max_{\substack{\graph \subseteq \searchdg \cr \graph \in \order}} P( \graph \mid \obs) \propto \prod_{i=1}^{n} \max_{\substack{\Pa_i \subseteq \searchdgpa^i \cr \Pa_i \in \order} } \score(X_i, \Pa_i \mid \obs) \,
\end{equation}
To compute the terms on the right we again follow the steps detailed in \aref{sumtableapp} using the Hasse power set representation of the permissible parent set of each node and propagating the maximum score down the power set following \alref{parentmaxalg} of \aref{sumtableapp}:
\begin{equation}\label{orderScoresdgmaxtable}
\Sordmax_{\searchdg}(\order \mid \obs) \propto \prod_{i=1}^{n} \maxscore^{i}_{f(\substack{\searchdgpa^i \in \order})} \, , \qquad \maxscore^{i}_{f(\substack{\searchdgpa^i \in \order})} = \max_{\substack{\Pa_i \subseteq \searchdgpa^i \cr \Pa_i \in \order} } \score(X_i, \Pa_i \mid \obs)
\end{equation}

\subsubsection{Stochastic search}

Finding the order with the highest $\Sordmax$ directly provides a MAP DAG.  A stochastic search based on the order MCMC scheme with score $\Sordmax_{\searchdg}(\order \mid \obs)$ can tackle the problem. Running the scheme, we keep track of the highest scoring order, and hence the highest scoring DAG, encountered.  The convergence time to sample orders from a distribution proportional to $\Sordmax_{\searchdg}(\order \mid \obs)$ is again expected to be $O(n^2\log(n))$.

To perform adaptive tempering and speed up discovery of a MAP DAG, we can transform the score by raising it to the power of $\gamma$ and employ $\Sordmax_{\searchdg}(\order \mid \obs)^{\gamma}$.  This transformation smooths ($\gamma<1$) or amplifies ($\gamma>1$) the score landscape and the value of $\gamma$ can be tuned adaptively depending on the acceptance probability of the MCMC moves while running the algorithm. To effectively explore local neighbourhoods, the target acceptance of swaps may scale $\propto \frac{1}{n}$.  Alternatively, simulated annealing can be performed by sending $\gamma\to\infty$.

\subsubsection{Greedy search}

The order-based scheme can also be adapted to perform a greedy search \citep{tk05}.  For example we score all possible $(n-1)$ local transpositions of adjacent nodes in $O(n)$ and select the highest scoring at each step.  Since it takes $O(n^2)$ steps to be able to reach any order with this move, we would expect $O(n^3)$ complexity to find each local maximum.  For the global swap of two random nodes, scoring the neighbourhood itself is $O(n^3)$ so that the $O(n)$ to traverse the space makes this move more expensive than local transpositions.  Local transpositions would therefore be generally preferable for greedy search, although global swaps may be useful to escape local maxima.

The new node relocation move of moving a single node at a time (\fref{ordermovepic}) requires only $O(n^2)$ to score all the possible new placements of all nodes.  With $O(n)$ steps to move between any pair of DAGs, we are again looking at $O(n^3)$ complexity for each search.  The new move also contains all local transpositions in its neighbourhood and provides a complementary alternative to a greedy search scheme purely based on local transpositions.

\subsection{Iteratively improving the search space}

Extending the search space, so that each node may have an additional parent outside the permissible set, allows us to discover edges which improve the DAG score, or edges with a high posterior weight (those which occur in a large fraction of sampled DAGs), which were previously excluded from the core search space defined by $\searchdg$.  Incorporating these edges into the core search space, we can iteratively improve the search space.  This is analogous to the iterative updating of \cite{fnp99}, but adapted to the order-based setting.

Starting with the original core search space $\searchdg_0 = \searchdg$ we expand to allow one additional parent, search for the maximally scoring DAG in that space and convert it to the CPDAG $\graph^{*}_0$.  For the next core search space $\searchdg_1$ we take the union of the edges in $\searchdg$ and $\graph^{\star}_0$, expand and search for the highest scoring CPDAG $\graph^{*}_1$ in the new space.  We iteratively repeat this procedure
\begin{equation}
\searchdg_{i+1} = \searchdg \cup \graph^{*}_i
\end{equation}
until no higher scoring CPDAG is uncovered and the last $\graph^{*}_i$ is entirely included in the core search space ($\searchdg_i =  \searchdg \cup \graph^{*}_i$).

\begin{figure}[t]
  \centering
  \includegraphics[width=0.85\textwidth]{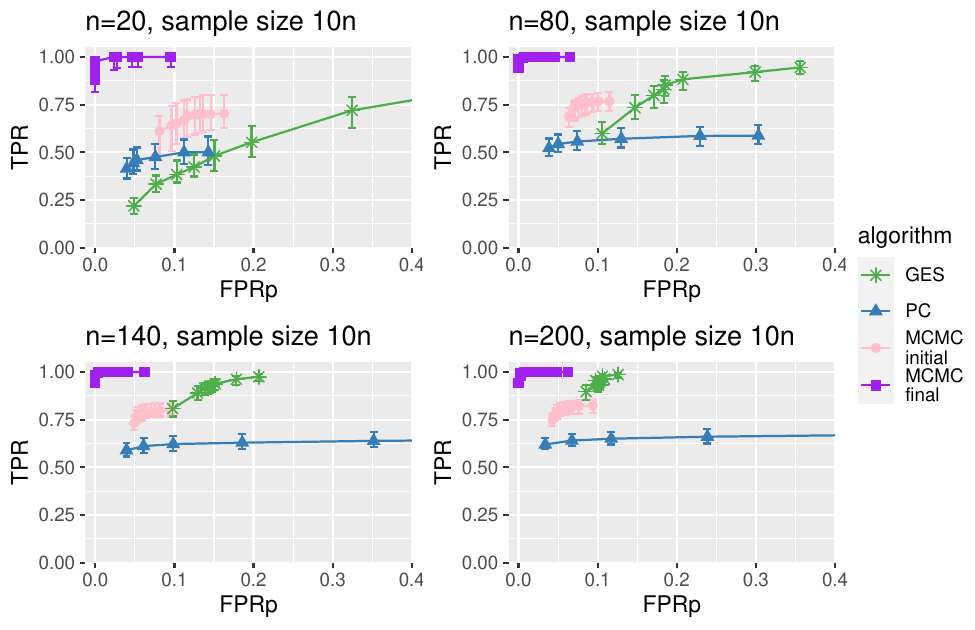}
  \caption{The performance in recovering the underlying DAG skeleton of our MCMC scheme (purple squares for different posterior thresholds) after converging to a core search space which contains the maximally scoring DAG encountered, compared to the PC algorithm (blue triangles for different significance levels) and GES (green stars for different likelihood penalisations).  For completeness we include the results of our MCMC scheme when forced to use the expanded initial search skeleton from the PC algorithm (pink circles).}
  \label{fig:ROC10n}
\end{figure}

\subsection{Performance} \label{sec:performance}

Detailed simulations illustrating the performance of our structure learning and sampling method are in \aref{app:simulations}. \fref{fig:ROC10n} highlights the improvement that iteratively updating the search space and then sampling from it has over alternatives which also apply to the more challenging scenarios, in terms of network size and density, that our method can handle.

In the main text we restrict our attention to the leading-order theoretical complexity of the computational tasks. The actual runtime in practice will additionally depend on the implementation details. For completeness we explore the typical runtimes of the \BiDAG implementation in \aref{app:simulations}.

\section{Partition-based unbiased DAG sampling on a fixed search space} \label{sec:partition}

To obtain an unbiased sample of DAGs we can adapt the order-based sampling scheme with a reduced search space (\sref{sec:order}) to work with partition MCMC \citep{km17} instead of order MCMC.

Partition MCMC requires the nodes of a DAG to be assigned to a labelled ordered partition $\Party=(\party,\permy_\party)$ consisting of a partition $\party$ of the $n$ nodes and a permutation $\permy_\party$ where the nodes within each partition element, called a \emph{block}, take ascending order.  The representation of each DAG as an ordered partition is unique unlike for the simpler order representation which has a bias towards DAGs belonging to multiple orders.  The assignment can be performed by recursively tracking nodes with no incoming edges, called outpoints \citep{robinson70,robinson73}.  In \fref{fig:dagsorderpartition} the outpoints are nodes 1 and 5 which are placed in the rightmost block. Removing these nodes and their outgoing edges, nodes 3 and 4 become outpoints and go into the second block; finally node 2 fills the remaining block.  The partition is $\lambda=[1,2,2]$ with permutation $\permy_\party=2,3,4,1,5$.

When reversing the process and building a DAG recursively, the outpoints at each stage must be connected to from the next set of outpoints.  Each node in any block must have at least one incoming edge from nodes in the adjacent block to the right, if there is one.  For example, node 2 in any DAG compatible with the partition in \fref{fig:dagsorderpartition}(c) must have an edge from either node 3 or node 4 (or both).  Additional edges may only come from nodes in blocks further right.

There are 12 possible incoming edge combinations for node 2, three for each of node 3 and 4, for a total of 108 DAGs compatible with the labelled ordered partition of the DAG in \fref{fig:dagsorderpartition}(c).  In partition MCMC we assign the sum of the scores of all these DAGs to the partition.  Below we describe an efficient implementation when the search space is restricted.

\subsection{Scoring partitions on a restricted search space} 

The posterior probability of a labelled partition is the sum of posterior probabilities of DAGs within the search space compatible with the partition
\begin{equation} \label{partitionscoreeqdg}
P_{\searchdg}(\Party \mid \obs) = \sum_{\substack{\graph \subseteq \searchdg \cr \graph \in \Party}} P(\graph \mid \obs) 
	\propto \prod_{i=1}^{n} \sum_{\substack{\Pa_i \subseteq \searchdgpa^i \cr \Pa_i \in \Party}} \score(X_i, \Pa_i \mid \obs)
\end{equation}
where the restriction on parents sets induced by the partition is that they must contain at least one node from the adjacent block to the right.  To evaluate the sums in \eref{partitionscoreeqdg} for each subset of banned nodes (belonging to the same block or further left) we need to keep track of the subset of \emph{needed} nodes belonging to the block immediately to the right to ensure at least one is in the parent set. Analogously to the order case, we use a one-to-one mapping for indexing parent sets. With $K$ permissible parents for node $i$ we have $3^{K}$ possible subset pairs for which we use the ternary mapping:
\begin{equation} \label{ternarymappingeq}
g(\setVarexample,\setVarexamplet)=\sum_{j=1}^{K}I(\searchdgpal^{i}_j \in \setVarexample)3^{j-1} + 2\sum_{j=1}^{K}I(\searchdgpal^{i}_j \in \setVarexamplet)3^{j-1}
\end{equation}
with $\setVarexample$ representing the permissible parents, and $\setVarexamplet$ those of which at least one must be present.

We detail the algorithmic steps to compute these sums efficiently in \aref{sumtableapp} and \alref{partparentsumalg} with a time complexity of $O(K^23^K)$ and space complexity of $O(3^K)$.  Unsurprisingly the restriction encoded by partitions to remove the bias of order MCMC also increases the computational cost of building the lookup tables for the partition MCMC sampler.  However, once the score table is built, computing the score of any partition from \eref{partitionscoreeqdg} reduces to
\begin{equation}\label{partitionscoreeqdgtable}
P_{\searchdg}(\Party \mid \obs) \propto \prod_{i=1}^{n} \sumscoret^{i}_{g(\substack{\searchdgpa^i \in \Party, \, \searchdgpa^i \in \party_i})} \, , \qquad  \sumscoret^{i}_{g(\substack{\searchdgpa^i \in \Party, \, \searchdgpa^i \in \party_i})} = \sum_{\substack{\Pa_i \subseteq \searchdgpa^i \cr \Pa_i \in \Party}} \score(X_i, \Pa_i \mid \obs)
\end{equation}
where $\party_i$ represents the block containing node $i$.  The score of the partition can be evaluated in $O(Kn)$ from the lookup tables.

\subsection{Partition MCMC moves}

The simplest move in the partition space consists of splitting blocks, or joining adjacent ones.  Proposing such a move from $\Party$ to $\Party'$ and accepting with probability
\begin{equation} \label{partitionacceptratio}
\rho =  \min \left \{ 1 , \frac{\neibr{\Party} P_{\searchdg}( \Party' \vert \obs) }{\neibr{\Party'} P_{\searchdg}( \Party \vert \obs)} 
		\right \} \, ,
\end{equation}		
while accounting for the neighbourhood sizes \citep[following][]{km17} is sufficient to sample partitions proportionally to their posterior probability in the search space.  Only nodes whose block changes between the current and proposed partition need to be rescored.  Although blocks can get to a size $O(n)$, on average they contain around $1.5$ nodes \citep{km15} so we would expect $O(1)$ for this move. Analogously to order MCMC (\ref{sec:dagSampling}), partition MCMC does not directly produce a chain of DAGs, but rather a chain of partitions from the posterior distribution $P_{\searchdg}( \Party \vert \obs)$. Conditionally on a sampled partition we can then sample a compatible DAG.  

To speed up convergence, \cite{km17} included additional permutation moves, either between two nodes in adjacent blocks, again requiring rescoring only nodes whose block changes in the proposed partition, or between any two nodes in different blocks also requiring the rescoring of all nodes inbetween.  We would typically expect $O(1)$ for the local swapping of nodes and $O(n)$ for the global swapping.  The global swap is picked with a probability $\propto \frac{1}{n}$ to contain the average complexity.

\begin{figure}[t]
  \centering
  \includegraphics[width=0.3\textwidth]{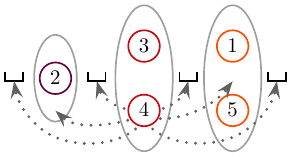}
  \caption{For any randomly chosen node, here 4, we score the entire neighbourhood of new positions by sequentially moving the node into different blocks or the gaps inbetween.  The new placement is sampled proportionally to the scores of the different labelled ordered partitions in the neighbourhood.}
  \label{partitionmovepic}
\end{figure}

Another possible move is to select a single node and move it elsewhere in the partition, or as a new block.  Analogously to the node relocation move introduced in \sref{sec:Gibbsmove} we can score the entire neighbourhood for any node selected at random by sequentially moving it through the partition, as in \fref{partitionmovepic}.  Since each other node has its needed or banned parent sets essentially affected twice, scoring the neighbourhood takes $O(n)$.  We always accept this move as it is sampled directly from the neighbourhood (which includes the starting partition), further aiding convergence.  This move is also selected with a probability $\propto \frac{1}{n}$.

\subsection{Expanding the search space}

When expanding the search space, for each node we simply create further summed score tables each including one other node from outside the search space as an additional parent. Additional parents, when located in the adjacent block to the right, will fulfil the requirement to have at least one parent from that set of nodes.  The space and time complexity increase by a factor of $n$ when building these tables. For the MCMC moves, the time complexity can increase by a factor $O(n)$ but the typical increase is again $O(K)$ on average.

\section{Discussion}
\label{discussion} 

In this work we presented a novel and computationally efficient approach for learning and sampling the DAGs underlying high dimensional Bayesian network models. Two main original features are worth highlighting:

First, the computational efficiency we gain by observing that every score quantity needed for the MCMC scheme can be effectively precomputed and stored in lookup tables.  This goes beyond the common strategy in DAG inference of simply storing the scores of individual parent sets by additionally storing all sums of parent set scores.  This allows us to sink the time and space complexity class of methods which reduce the search space by collectively scoring groups of DAGs, as in order-based approaches.  Specifically, we reduce the main complexity bottleneck of $n^K$ to just $2^K$ or $3^K$ providing a massive computational advantage for larger and denser networks.  Order and partition MCMC constitute the building blocks of our procedure with the added benefit that each step in the chain now takes minimal computational time.

Second, the improved accuracy in the network structure inference, achieved by means of an iterative expansion of the search space beyond the preliminary skeleton obtained through constraint-based methods. In fact the pre-defined search space may not include DAGs corresponding to the mode of the posterior distribution, so that hybrid methods can heavily benefit from the additional flexibility.  The simulation studies (\aref{app:simulations}) extensively demonstrate the improved performance we can achieve with respect to current mainstream approaches in terms of the accuracy of the reconstructed network. In special cases where we expect sparse graphs traditional methods like PC and GES will achieve similar accuracies in much less time and may be preferred. As the networks grow denser however, their accuracy deteriorates, while our hybrid MCMC algorithm continues to provide reliable results, even for relatively large networks of practical interest for real data. All methods will suffer with very large networks; we need to be cautious in trusting edges returned by traditional methods, while our algorithm becomes computationally more demanding.

When iteratively updating the search space, we include edges ensuring that the highest scoring DAG found at each stage belongs to the core search space for the next iteration. Alternatively we could update the search space by incorporating edges with a certain posterior probability when sampling over the extended space at each iteration. The order-based sample is of course biased, but the additional restriction in the combinatorics of partition MCMC, which ensures a unique representation of each DAG, increases the complexity of building the necessary lookup tables.  For denser networks it may be preferable to pursue bias removal only at later iterations, once the search space has already converged under order.  Finding the highest scoring DAG or sampling with order MCMC share the same complexity. We chose to update the search space based on the highest scoring DAG since order MCMC may find a maximal score faster than sampling, thanks to the possibility of tempering. 

The freedom to add edges beyond a pre-defined skeleton, allows for the correction of errors where edges may be missed.  The iterative approach is, aside from stochastic fluctuations in the search or sampling, mainly deterministic.  However, since we only consider the addition of a single parent at a time for each node, the algorithm may not pick up missing correlated edges, which would only improve the score if added at the same time.  Allowing for the concomitant addition of edge pairs increases the overall space complexity by a factor $n$ which can be computationally prohibitive. On the other hand we could view the search space itself, or the lists of permitted parents, as a random variable and implement a stochastic updating scheme. Especially for sparser graphs, such a scheme may be effective at extending the posterior sample outside of a fixed search space.

As the initial core search space we adopted the undirected skeleton obtained from the PC algorithm, without accounting for any orientations. The iterative steps of building the score tables have exponential complexity in the number of parents.  In the case of nodes with many children, which will be included as potential parents, ignoring the direction will lead to increased costs in building the lookup tables. In certain cases it may therefore be convenient to limit permissible parent sets of particular nodes to those compatible with directed or undirected edges in the CPDAG learned through the PC algorithm.

Despite our focus on taking the skeleton given by the PC algorithm as the initial core search space, our approach is agnostic to the method used to define the starting point, although obviously performance will improve the closer the initial search space is to the target space containing the bulk of the posterior distribution. If relevant edges are missing in the initial search space, our algorithm can add them though it may take a few iterations to do so.  False positive edges in the search space do not affect the MCMC search, but do increase the time and space needed for computing the lookup tables.  In our simulations, the PC algorithm was quite conservative, even when relaxing the significance threshold, missing many edges but introducing few false positives. Due to the large number of missing edges, improving the search space tended to require quite a few iterations, which were however reasonably fast. 

Defining the initial core search space by GES would include more of the important edges to start with, but also many false positives. As a consequence the algorithm would potentially require fewer steps to improve the search space, at the expense of higher computational cost of each step. In the context of GES, the number of false positives is sensitive to the penalisation parameter in the score, so ideally we should optimally tune it if using GES to define the initial search space.  Order-based conditional independence tests \citep{ru18} also offer another option.  For Gaussian models, the Markov random field or conditional independence graph defined by the precision matrix \citep[as used for example in][]{nhm18} is also a possibility.  Theoretically the conditional independence graph should contain all edges present in the PC algorithm skeleton, potentially including more true positive edges, while most likely also introducing additional false positives. In principle one may even combine search spaces from different approaches. Given a consistent initial search space, and the recent advances in proving consistency in hybrid methods \citep{nhm18}, an interesting open question is whether order-based searches as developed here can be proven or modified to be consistent.

Interesting directions may also come from the ILP method of \cite{cussens11} and \cite{cussensetal17}, if the solver manages to complete and the number of parents in the maximally scoring DAG is less than the low limit needed for their input score tables.  By expanding such a DAG appropriately, we may obtain a good starting point for the full sampling.  Conversely, the final search space obtained by our search could be an interesting input for the ILP, or may be determined by combining elements of both approaches.  Similarly one may investigate whether one can modify dynamic programming approaches for exhaustively searching orders \citep{ks04,sm06,em07,htw16} to work on restricted search spaces and be efficient enough to replace the MCMC search.

Regardless of how we define the initial search space, or how we discover the maximal DAG, our hybrid scheme is the only one capable of efficiently sampling larger and denser graphs.  Sampling from the posterior distribution not only improves structure learning, but is vital for understanding the uncertainty in the graph structure itself.  To achieve robust inference we need to account for the structure uncertainty in analyses further downstream \citep{kuipersetal18}, for example for causal interpretations and in the estimation of intervention effects \citep{moffaetal17}.

\section*{Acknowledgements}

The authors would like to thank Markus Kalisch and Niko Beerenwinkel for useful comments and discussions.

\clearpage

\appendix

\section*{Supplementary Material}
\setcounter{section}{0} 

\renewcommand\thefigure{S\arabic{figure}}
\setcounter{figure}{0} 
\renewcommand\thetable{S\arabic{table}}
\setcounter{table}{0} 
\renewcommand\thealgorithm{S\arabic{algorithm}}
\setcounter{algorithm}{0} 

\section{Algorithmic details for computing summed score tables}
\label{sumtableapp}

\begin{table}
\begin{center}
\begin{tabular}{l|l c l|l}
Parent nodes & Node score & & Banned parents & Summed node score\\
\cline{1-2} \cline{4-5}
$\emptyset$ & $\score^{i}_{0} =  \score(X_i, \{\emptyset\} \mid \obs)$ & & $\emptyset$ & $\sumscore^{i}_{7} =  \sum_{j=0}^{7}\score^{i}_{j}$ \\
$\searchdgpal^i_1$ & $\score^{i}_{1} =  \score(X_i, \{\searchdgpal^i_1\}\mid \obs)$ & & $\searchdgpal^i_1$ & $\sumscore^{i}_{6} =  \score^{i}_{0}+\score^{i}_{2}+\score^{i}_{4}+\score^{i}_{6}$ \\
$\searchdgpal^i_2$ & $\score^{i}_{2} =  \score(X_i, \{\searchdgpal^i_2\} \mid \obs)$ & & $\searchdgpal^i_2$ & $\sumscore^{i}_{5} =  \score^{i}_{0}+\score^{i}_{1}+\score^{i}_{4}+\score^{i}_{5}$ \\
$\searchdgpal^i_3$ & $\score^{i}_{4} =  \score(X_i, \{\searchdgpal^i_3\} \mid \obs)$ & & $\searchdgpal^i_3$ & $\sumscore^{i}_{3} =  \score^{i}_{0}+\score^{i}_{1}+\score^{i}_{2}+\score^{i}_{3}$ \\
$\searchdgpal^i_1, \searchdgpal^i_2$ & $\score^{i}_{3} =  \score(X_i, \{\searchdgpal^i_1,\searchdgpal^i_2\} \mid \obs)$ & & $\searchdgpal^i_1, \searchdgpal^i_2$ & $\sumscore^{i}_{4} =  \score^{i}_{0}+\score^{i}_{4}$ \\
$\searchdgpal^i_1, \searchdgpal^i_3$ & $\score^{i}_{5} =  \score(X_i, \{\searchdgpal^i_1,\searchdgpal^i_3\} \mid \obs)$ & & $\searchdgpal^i_1, \searchdgpal^i_3$ & $\sumscore^{i}_{2} =  \score^{i}_{0}+\score^{i}_{2}$ \\
$\searchdgpal^i_2, \searchdgpal^i_3$ & $\score^{i}_{6} =  \score(X_i, \{\searchdgpal^i_2,\searchdgpal^i_3\} \mid \obs)$ & & $\searchdgpal^i_2, \searchdgpal^i_3$ & $\sumscore^{i}_{1} =  \score^{i}_{0}+\score^{i}_{1}$ \\
$\searchdgpal^i_1, \searchdgpal^i_2, \searchdgpal^i_3$ & $\score^{i}_{7} =  \score(X_i, \{\searchdgpal^i_1,\searchdgpal^i_2, \searchdgpal^i_3\} \mid \obs)$ & & $\searchdgpal^i_1, \searchdgpal^i_2, \searchdgpal^i_3$ & $\sumscore^{i}_{0} =  \score^{i}_{0}$ 
\end{tabular}
\end{center}
\caption{An example score table of a node with 3 permissible parents in the search space (left).  For each possible list of excluded parents, we also create a second table (right) containing the sum of scores of all subsets of remaining parents.}
\label{scoretableexample}
\end{table}

Given the scores of all permissible parent sets of a node (\tref{scoretableexample}, left), we detail how to compute the summed score (\tref{scoretableexample}, right).  The first column indicates which nodes are banned as parents and the second column reports the sum of scores over all parent subsets excluding those nodes.  For the indexing of the sums we negate the indicator function:
\begin{equation}
\tf(\setVarexample)=\sum_{j=1}^{K}I(\searchdgpal^{i}_j \notin \setVarexample)2^{j-1} = 2^{K}-f(\setVarexample)-1
\end{equation}

\subsection{Power set representation}

\begin{figure}[t]
  \centering
  \includegraphics[width=0.65\textwidth]{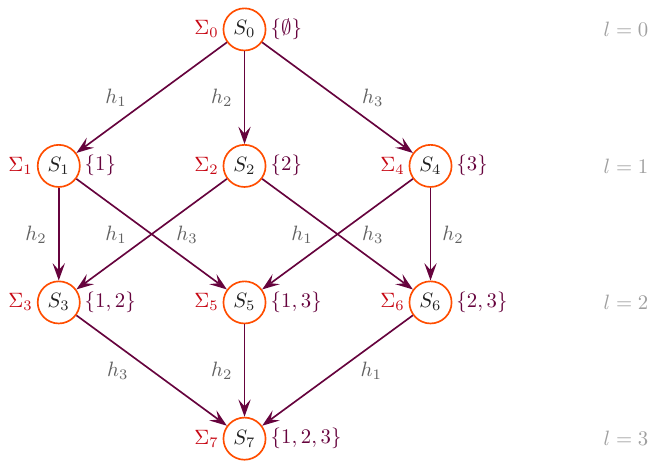}
  \caption{In the Hasse diagram the permissible parent subsets can be arranged by their size and connected as a diagram where parents are added along each directed edge.  The subsets are indicated on the right side of each node.  Inside the nodes of the diagram are the scores of that particular parent subset.  To the left of each node we place the sum of scores of that node and all its ancestors in the diagram.  This sum encompass all possible subsets which exclude any members of the complement.}
  \label{possepic}
\end{figure}

A Hasse diagram (\fref{possepic}) visualises the power set of the permissible parents with layers ranked by the size of the parent subsets, and helps develop a strategy to efficiently evaluate the partial sums over parent subsets. Directed edges indicate the addition of another parent to each subset, while the corresponding scores of each parent subset are attached to the nodes in \fref{possepic}.  The advantage of the Hasse representation is that each element in the summed score table (right of \tref{scoretableexample}) is the sum of the scores of a node and all its ancestors in the diagram.  Power set representations have also been previously used for Bayesian network inference, for example by \cite{ks04} to sum over orders. 

\subsection{Score propagation}

To actually perform the sums we utilise the separation of the power set into $(K+1)$ layers of differently sized parent subsets and implement \alref{parentsumalg}.  The partial sums at each layer are propagated to their children in the network.  To avoid overcounting contributions from ancestors which are propagated along all $d!$ paths connecting nodes $d$ layers apart, we divide by the corresponding factorials to obtain the required sums. This division is separated over the layers by dividing by one of the factorial terms each time.  For each end layer there are a different number of ancestral paths to the nodes in previous layers leading to different correction factors, so we need to repeat the propagation $K$ times.   Building the summed score table for each variable has a complexity of $O(K^{2}2^{K})$: during each propagation, the value at each of the $2^{K}$ elements of the power set is created by adding its starting value with the values of its parents of which there can be at most $K$ giving a complexity of $O(K2^{K})$, while the propagation is repeated $K$ times.  

\begin{algorithm}[t]
\caption{Obtain the sum of scores of all parent subsets excluding banned nodes}\label{parentsumalg}
\begin{algorithmic}
\State \textbf{Input} The power set network of the $K$ permissible parents of variable $i$
\State \textbf{Input} The table of scores of each parent subset $\score^{i}_j$, $j=0,\ldots,(2^{K}-1)$
\State Label the network nodes $Y_{f(\setVarexample)}$ for each $\setVarexample$ in the power set
\State Initialise the node at layer 0: $\sumscore^{i}_{0} =  Y_{0} = \score^{i}_{0}$
\For{$l=1$ to $K$} \Comment{layer number}
\For{$m=0$ to $(l-1)$}
\State Initialise the value of all nodes in layer $(m+1)$:
\ForAll{$\{j \in \mbox{ layer } (m+1)\}$}
\State $Y_{j} = \score^{i}_{j}$
\EndFor
\State Add the value of nodes $Y_j$ at layer $m$, divided by $(l-m)$,
\State to all children in the power set network at layer $(m+1)$:
\ForAll{$\{j \in \mbox{ layer } m\}$}
\ForAll{$\{j' \mid Y_{j'} \mbox{ child of } Y_j\}$}
\State $Y_{j'} = Y_{j'}+\frac{Y_j}{(l-m)}$ \Comment{division accounts for overcounting}
\EndFor
\EndFor
\EndFor
\State Read off sum scores at layer $l$:
\ForAll{$\{j \in \mbox{ layer } l\}$}
\State $\sumscore^{i}_{j} = Y_{j}$
\EndFor
\EndFor
\State \textbf{return} Table of summed scores: $\sumscore^{i}_j$, $j=0,\ldots,(2^{K}-1)$
\end{algorithmic}
\end{algorithm}

\subsection{MAP DAG targetting}

When assigning the score of each order to be the maximum score of DAGs in the order, we do not need to worry about the overcounting and can propagate only once in $O(K2^{K})$, see \alref{parentmaxalg}.

\begin{algorithm}[h!t]
\caption{Obtain the maximal score among all parent subsets excluding banned nodes}\label{parentmaxalg}
\begin{algorithmic}
\State \textbf{Input} The power set network of the $K$ permissible parents of variable $i$
\State \textbf{Input} The table of scores of each parent subset $\score^{i}_j$, $j=0,\ldots,(2^{K}-1)$
\State Label the network nodes $Y_{f(\setVarexample)}$ for each $\setVarexample$ in the power set
\State Initialise all nodes: $Y_{j} = \score^{i}_{j}$, $j=0,\ldots,(2^{K}-1)$
\For{$l=1$ to $K$} \Comment{layer number}
\State Replace the value of nodes $Y_j$ at layer $l$ by the maximum of itself
\State and all its parents in the power set network at layer $(l-1)$:
\ForAll{$\{j \in \mbox{ layer } l\}$}
\ForAll{$\{j' \mid Y_{j'} \mbox{ parent of } Y_j\}$}
\State $Y_{j} = \max(Y_{j},Y_{j'})$
\EndFor
\EndFor
\EndFor
\State \textbf{return} Table of maximal scores: $\maxscore^{i}_{j}=Y_j$, $j=0,\ldots,(2^{K}-1)$
\end{algorithmic}
\end{algorithm}

\subsection{Restricted sums}

For the partition based sampling, we need to ensure that nodes receive at least one edge from the adjacent block. For the example with 3 permissible parents, there are the 8 values calculated in \tref{scoretableexample} where there was no restriction on enforcing the presence of a member of the needed parent subset (which we regard as the empty set).  Additionally there are the 19 combinations in \tref{partitionscoretableexample} where we index the sums with the ternary mapping of \eref{ternarymappingeq}.  We also define the mapping back to the banned parent set
\begin{equation} \label{ternarymappingbackeq}
\tg(j)= \setVarexample: g(\setVarexample,\setVarexamplet) = j 
\end{equation}
\begin{table}
\begin{center}
\begin{tabular}{l|l|l}
Banned parents & Needed parents & Summed node score\\
\cline{1-3}
$\emptyset$ & $\searchdgpal^i_1$ & $\sumscoret^{i}_{2} =  \score^{i}_{1}+\score^{i}_{3}+\score^{i}_{5}+\score^{i}_{7}$ \\
$\emptyset$ & $\searchdgpal^i_2$ & $\sumscoret^{i}_{6} =  \score^{i}_{2}+\score^{i}_{3}+\score^{i}_{6}+\score^{i}_{7}$ \\
$\emptyset$ & $\searchdgpal^i_3$ & $\sumscoret^{i}_{18} =  \score^{i}_{4}+\score^{i}_{5}+\score^{i}_{6}+\score^{i}_{7}$ \\
$\emptyset$ & $\searchdgpal^i_1,\searchdgpal^i_2$ & $\sumscoret^{i}_{8} =  \score^{i}_{1}+\score^{i}_{2}+\score^{i}_{3}+\score^{i}_{5}+\score^{i}_{6}+\score^{i}_{7}$ \\
$\emptyset$ & $\searchdgpal^i_1,\searchdgpal^i_3$ & $\sumscoret^{i}_{20} =  \score^{i}_{1}+\score^{i}_{3}+\score^{i}_{4}+\score^{i}_{5}+\score^{i}_{6}+\score^{i}_{7}$ \\
$\emptyset$ & $\searchdgpal^i_2,\searchdgpal^i_3$ & $\sumscoret^{i}_{24} =  \score^{i}_{2}+\score^{i}_{3}+\score^{i}_{4}+\score^{i}_{5}+\score^{i}_{6}+\score^{i}_{7}$ \\
$\emptyset$ & $\searchdgpal^i_1,\searchdgpal^i_2,\searchdgpal^i_3$ & $\sumscoret^{i}_{26} =  \score^{i}_{1}+\score^{i}_{2}+\score^{i}_{3}+\score^{i}_{4}+\score^{i}_{5}+\score^{i}_{6}+\score^{i}_{7}$ \\
$\searchdgpal^i_1$ & $\searchdgpal^i_2$ & $\sumscoret^{i}_{7} =  \score^{i}_{2}+\score^{i}_{6}$ \\
$\searchdgpal^i_1$ & $\searchdgpal^i_3$ & $\sumscoret^{i}_{19} =  \score^{i}_{4}+\score^{i}_{6}$ \\
$\searchdgpal^i_1$ & $\searchdgpal^i_2, \searchdgpal^i_3$ & $\sumscoret^{i}_{25} =  \score^{i}_{2}+\score^{i}_{4}+\score^{i}_{6}$ \\
$\searchdgpal^i_2$ & $\searchdgpal^i_1$ & $\sumscoret^{i}_{5} =  \score^{i}_{1}+\score^{i}_{5}$ \\
$\searchdgpal^i_2$ & $\searchdgpal^i_3$ & $\sumscoret^{i}_{21} =  \score^{i}_{4}+\score^{i}_{5}$ \\
$\searchdgpal^i_2$ & $\searchdgpal^i_1, \searchdgpal^i_3$ & $\sumscoret^{i}_{23} =  \score^{i}_{1}+\score^{i}_{4}+\score^{i}_{5}$ \\
$\searchdgpal^i_3$ & $\searchdgpal^i_1$ & $\sumscoret^{i}_{11} =  \score^{i}_{1}+\score^{i}_{3}$ \\
$\searchdgpal^i_3$ & $\searchdgpal^i_2$ & $\sumscoret^{i}_{15} =  \score^{i}_{2}+\score^{i}_{3}$ \\
$\searchdgpal^i_3$ & $\searchdgpal^i_1, \searchdgpal^i_2$ & $\sumscoret^{i}_{17} =  \score^{i}_{1}+\score^{i}_{2}+\score^{i}_{3}$ \\
$\searchdgpal^i_1, \searchdgpal^i_2$ & $\searchdgpal^i_3$ & $\sumscoret^{i}_{22} =  \score^{i}_{4}$ \\
$\searchdgpal^i_1, \searchdgpal^i_3$ & $\searchdgpal^i_2$ & $\sumscoret^{i}_{16} =  \score^{i}_{2}$ \\
$\searchdgpal^i_2, \searchdgpal^i_3$ & $\searchdgpal^i_1$ & $\sumscoret^{i}_{14} =  \score^{i}_{1}$
\end{tabular}
\end{center}
\caption{An example sum score table for each possible list of excluded parents, where at least one member of the needed parents subset must be included.}
\label{partitionscoretableexample}
\end{table}

\begin{figure}[t]
  \centering
  \includegraphics[width=\textwidth]{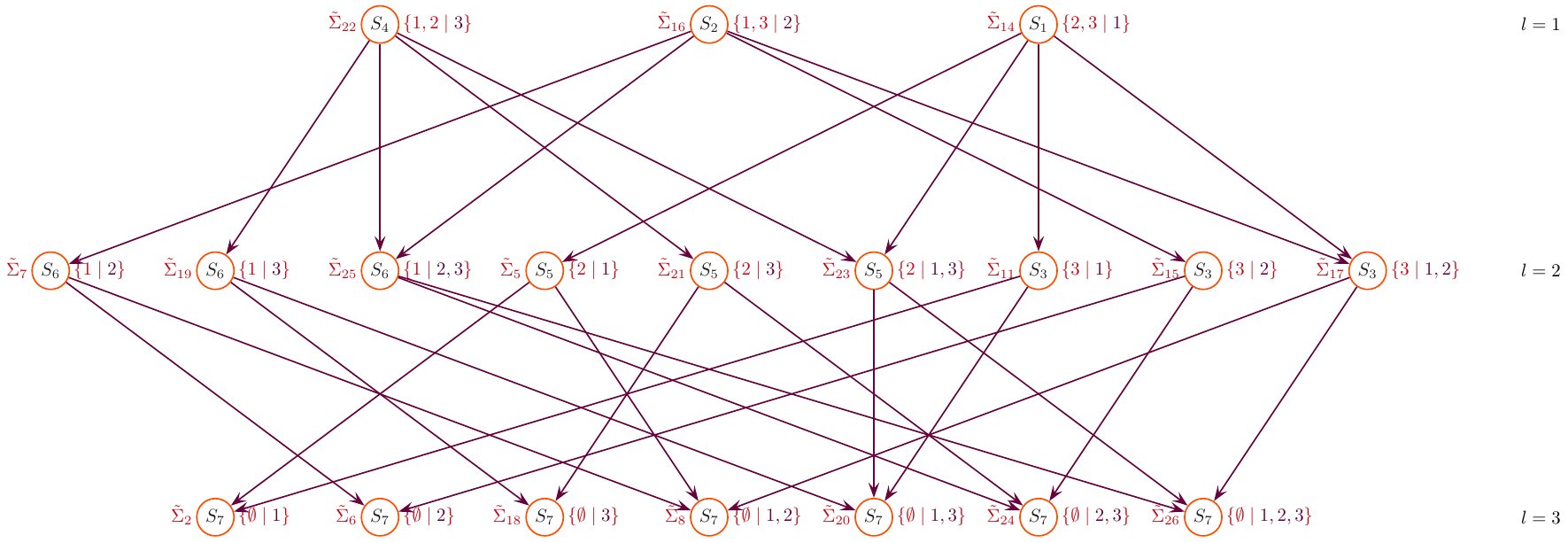}
  \caption{The banned parent subsets can be arranged by their size and expanded to include all needed parent subsets in the complement.  Inside the nodes of the network are the scores of the complement of the banned parent subset, and both the banned and needed parents subsets are indicated on the side of each node  The nodes are connected as a network where parents are deleted from the banned parent subsets or moved into the needed parent subsets.  The sum of all scores which do not involve certain banned parents but do include at least one member of the needed parent subset is simply the sum of scores associated with a node and all its ancestors in the network.}
  \label{partpossepic}
\end{figure}

Again we build a network representation of the possibilities by replicating each node in the power set representation according to the number of choices of possible needed parent subsets in the complement of the banned parent subsets.  We rank the nodes by the size of the banned node subsets as in \fref{partpossepic}, and assign to the nodes the score corresponding to the complement of the banned node subset. The connections in the network represent either removing an element from the banned parent subset, or moving it to the needed parent subset.  For any $j$ such that $\searchdgpa^i_j$ is in the banned parent subset there is then an edge to the node indexed by $3^{j-1}$ more and the node indexed by $3^{j-1}$ less using the ternary mapping of \eref{ternarymappingeq}. 

The sums in \tref{partitionscoretableexample} are the sums of the scores of each node in the network in \fref{partpossepic} and its ancestors.  To compute these sums efficiently we again propagate through the network using \alref{partparentsumalg} whose complexity is $O(K^23^K)$.  

\begin{algorithm}[h!t]
\caption{Obtain the sum of scores of all parent sets excluding all banned nodes but including at least one member of needed nodes}\label{partparentsumalg}
\begin{algorithmic}
\State \textbf{Input} The network of the banned and needed parent subsets of variable $i$ from the $K$ permissible parents
\State \textbf{Input} The table of scores of each parent set $\score^{i}_j$, $j=0,\ldots,(2^{K}-1)$
\State \textbf{Input} The table of summed scores for each banned parent subset $\sumscore^{i}_j$, $j=0,\ldots,(2^{K}-1)$
\State Label the network nodes $Y_{g(\setVarexample,\setVarexamplet)}$
\State Initialise the restricted summed scores for empty needed nodes:
\For{$j=0$ to $(2^{K}-1)$}
\State $\sumscoret^{i}_{g(f^{-1}(j),\emptyset)} = \sumscore^{i}_{j}$
\EndFor
\State Initialise the nodes at layer 1:
\ForAll{$\{j \in \mbox{ layer } 1\}$}
\State $\sumscoret^{i}_{j} = Y_{j} = \score^{i}_{\tf(\tg(j))}$
\EndFor
\For{$l=2$ to $K$}
\For{$m=1$ to $(l-1)$}
\State Initialise the value of all nodes in layer $(m+1)$:
\ForAll{$\{j \in \mbox{ layer } (m+1)\}$}
\State $Y_{j} = \score^{i}_{\tf(\tg(j))}$
\EndFor
\State Add the value of nodes $Y_j$ at layer $m$, divided by $(l-m)$,
\State to all children in the network at layer $(m+1)$:
\ForAll{$\{j \in \mbox{ layer } m\}$}
\ForAll{$\{j' \mid Y_{j'} \mbox{ child of } Y_j\}$}
\State $Y_{j'} = Y_{j'}+\frac{Y_j}{(l-m)}$
\EndFor
\EndFor
\EndFor
\State Read off restricted sum scores at layer $l$:
\ForAll{$\{j \in \mbox{ layer } l\}$}
\State $\sumscoret^{i}_{j} = Y_{j}$
\EndFor
\EndFor
\State \textbf{return} Table of restricted summed scores: $\sumscoret^{i}_j$, $j=0,\ldots,(3^{K}-1)$
\end{algorithmic}
\end{algorithm}

\section{Simulation studies}
\label{app:simulations}

To examine the performance and convergence of our method, we performed a simulation study for 4 network sizes $n\in\{20, 80, 140, 200\}$ and 2 sample sizes $N\in\{2n, 10n\}$.   For each combination of network and sample sizes, 100 random graphs were generated using the \texttt{randDAG} function from the \pkg{pcalg} package using the default Erd\H{o}s-R\'enyi model with an average parent set size of $2$.  The strengths of the edges were sampled uniformly in the range $[0.4, 2]$. Continuous data was generated in topological order following a linear structural equation model: $X_i = \sum_{X_j \in \Pa_i}\beta_{ji}X_j + \epsilon$ where $\beta\sim U(0.4, 2)$ are the edge strengths and $\epsilon\sim N(0,1)$ is Gaussian noise, and then standardised.  Although the average number of parents is 2, the determining factor for the runtime is the maximal number of parents.  With 20 nodes the maximal number of parents is 6 on average, rising to 8 on average with 80 nodes (with more than 5\% of cases having 10 parents of more) and further increases for larger networks.

\begin{figure}[t]
  \centering
  \includegraphics[width=0.75\textwidth]{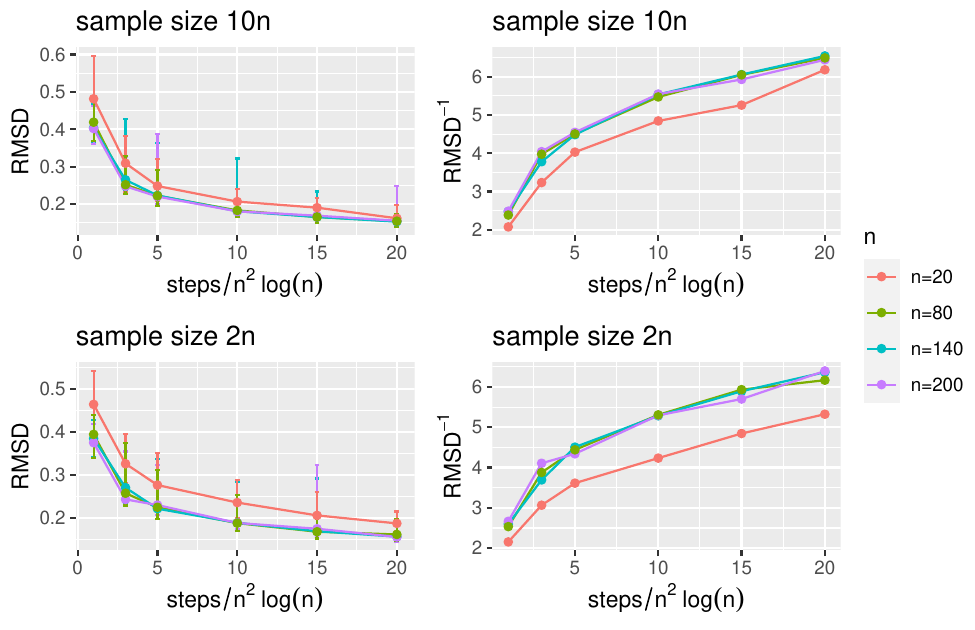}
  \caption{The root mean square difference (RMSD) between edge probabilities from pairs of different runs for each simulation as the size of the network increases.  The runs are those from \fref{fig:R2convergence} where the correlation $\rho^2$ was displayed.}
  \label{fig:RMSDconvergence}
\end{figure}

Methods which impose a hard limit on the number of parents, as is the case with leading score-based schemes \citep{fk03,tk05,gh08,km17} including ILP solvers \citep{cussens11,cussensetal17}, scale in complexity as $O(n^{K+1})$ and simply do not scale to our simulation setting. Order-based schemes for MAP DAG discovery, can prune the sets of permissible parents to reduce this computational burden \citep{fk03,tk05}.  A recent implementation \citep{scanagatta2015learning} can only handle categorical variables, so we compare to that approach in \sref{catsim}.

For continuous data we therefore compare only to GES \citep{chickering02}, a greedy structure-based search, and the PC algorithm \citep{bk:sgs00}, a constraint-based method. The hybrid approach of \cite{tba06} of a greedy structure search in a constraint-based skeleton performs very similarly to the PC algorithm and is not included in the comparisons.

For our method we use the BGe score with priors $\alpha_\mu = 0.25$ and $\alpha_w = \alpha_\mu + n + 1$. The significance level for the initial PC algorithm is $\min(0.4, \frac{20}{n})$. The number of MCMC iterations for the iterative search is $\max(25000, 3.5n^2\log(n))$, and  $\max(25000, 5n^2\log(n))$ for subsequent sampling. Code to reproduce the continuous simulations is available at \url{github.com/jackkuipers/esslbn_sim}.

\begin{figure}[t]
  \centering
  \includegraphics[width=0.75\textwidth]{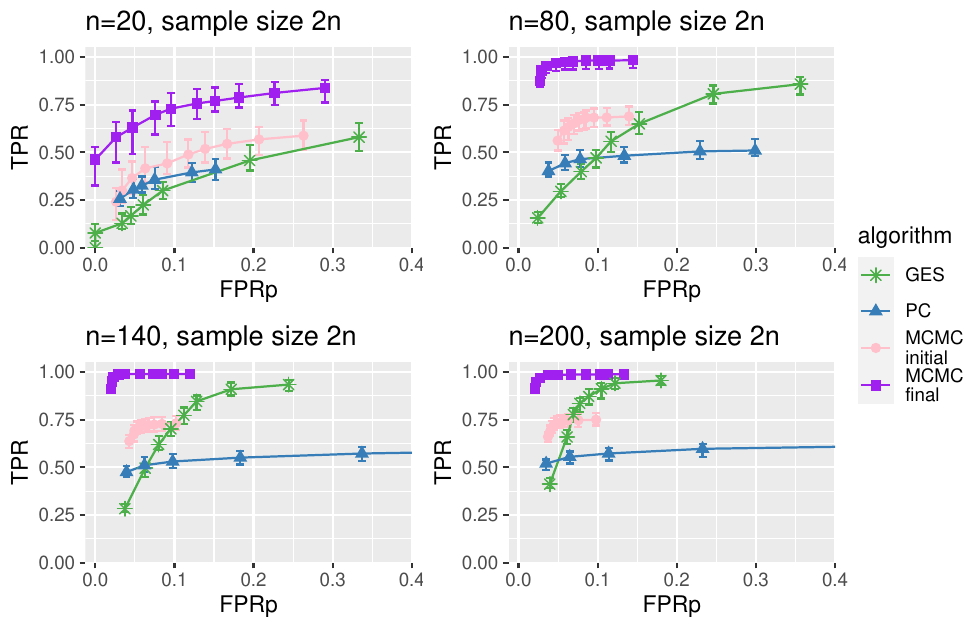}
  \caption{The performance in recovering the underlying DAG skeleton.  The graph is as \fref{fig:ROC10n} but with a smaller sample size of $N=2n$.}
  \label{fig:ROC2n}
\end{figure}
\begin{figure}[t]
  \centering
  \includegraphics[width=0.75\textwidth]{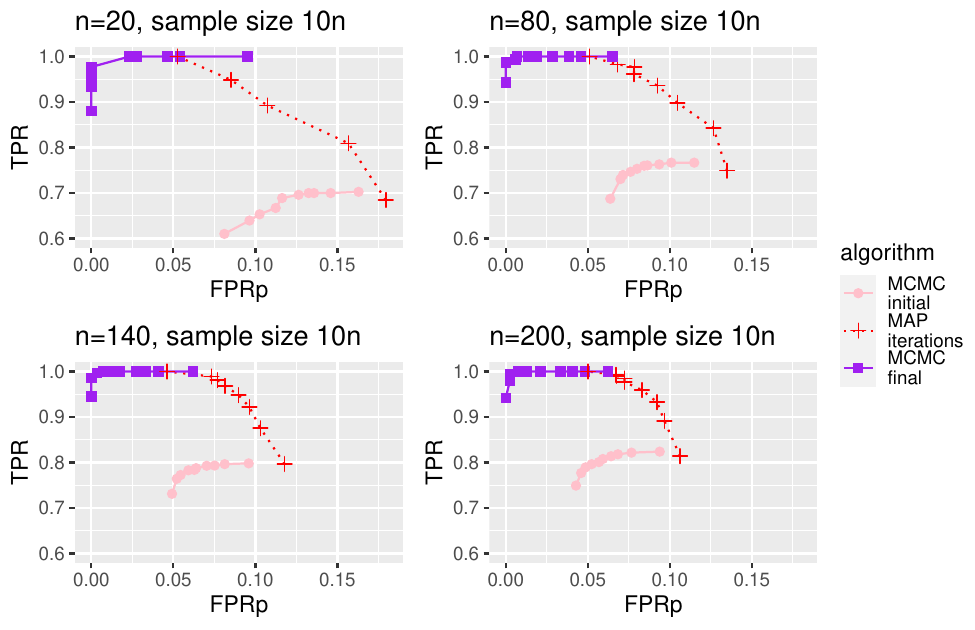}
  \caption{How the iterative search improves performance, using the setting of \fref{fig:ROC10n} as an example.  Starting from the search skeleton defined by the PC algorithm, expanded to include an possible additional parent (pink circles) we plot the highest scoring DAG discovered in that search space and during subsequent iterations (red plusses). When no better DAG is discovered, sampling from the final search space provides the purple squares.}
  \label{fig:iterations10n}
\end{figure}

\subsection{Skeleton inference}

To assess the performance we first considered the number of true positives (TP) and false positives (FP) in the undirected skeletons of the networks inferred and scaled them by the number of edges (P) in the true DAG
\begin{equation}
\mathrm{TPR} = \frac{\mathrm{TP}}{\mathrm{P}} \, \qquad \mathrm{FPRp} = \frac{\mathrm{FP}}{\mathrm{P}}
\end{equation}

We computed the median TPR along with the first and third quartiles and plotted (\frefs{fig:ROC10n} and \ref{fig:ROC2n}) against the median FPRp over the 100 realisations for our MCMC scheme for two search spaces: the initial search space defined by the PC algorithm skeleton, and expanded to include an additional parent; and the final search space which is improved iteratively until it contains the MAP DAG discovered. Also plotted are the results from GES \citep{chickering02} and the PC algorithm using Fisher's $z$ test for conditional independence \citep{bk:sgs00, art:KalischMCMB2012}.  The range of discrimination thresholds for plotting points in the ROC curves were:

\begin{itemize}
  \item penalisation parameter $\frac{\lambda}{\log (N)} \in \{1, 2,  5, 7,  9, 11, 15, 25\}$ for GES
  \item significance level $\alpha \in \{0.01, 0.05, 0.1, 0.2, 0.35, 0.45\}$ for the PC algorithm (the highest threshold may result in too many false positives to display in the plots)
  \item posterior probability $\rho \in \{0.2, 0.3, 0.4, 0.5, 0.6, 0. 7, 0.8, 0.9, 0.95, 0.99\}$ for MCMC 
\end{itemize}

With the initial search space, we see a distinct improvement with our MCMC scheme (pink circles in \frefs{fig:ROC10n} and \ref{fig:ROC2n}), while when we improve the search space iteratively we observe a strong advantage over the alternative methods (purple squares in \frefs{fig:ROC10n} and \ref{fig:ROC2n}) and approach perfect recovery of the skeleton for the larger sample size (\fref{fig:ROC10n}). 

In the simulations, increasing the significance level of the conditional independence tests of the PC algorithm does not really improve the recovery of true edges, while the additional false positives start to dramatically increase the algorithm's runtime.

\subsection{Iterative steps}
\label{moresimsapp}

To explore how the iterative search leads to an improvement in performance we keep track of the highest scoring DAG uncovered at each iteration, and used to update the core search space for the next iteration.  In \fref{fig:iterations10n}, we overlay these intermediate results on the MCMC lines of \fref{fig:ROC10n}.  Each iteration, and especially the earlier ones leads to an improvement in the search space allowing the MCMC search to find better DAGs which were previously excluded.  Finally, utilising the posterior probability of edges in the sample from the final search space, we can remove some of the false positive edges in the point estimate of the highest scoring DAG uncovered.

\begin{figure}[t]
  \centering
  \includegraphics[width=0.75\textwidth]{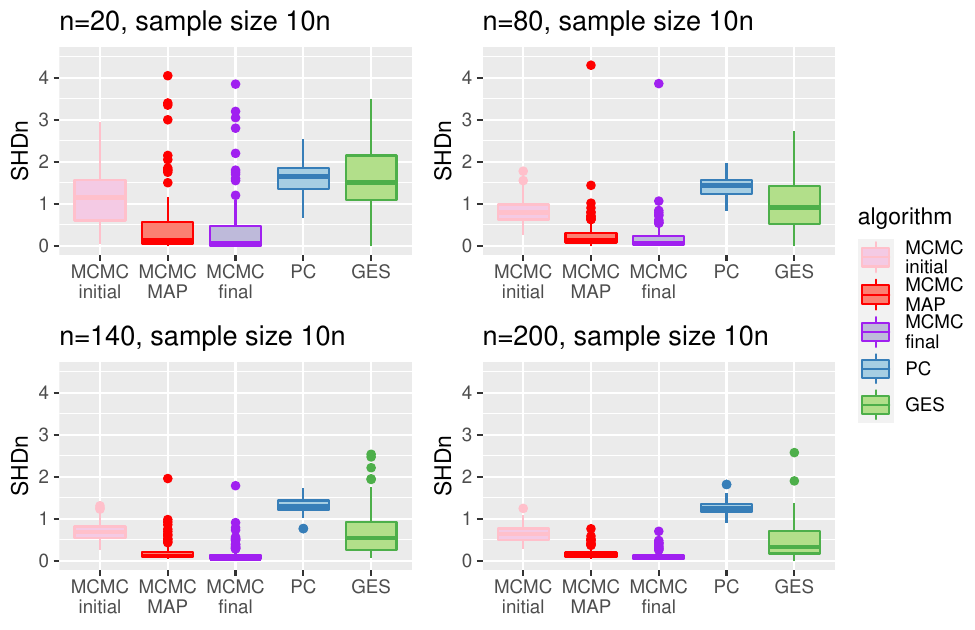}
  \caption{The performance in recovering the CPDAG measured by the structural Hamming distance (SHD) divided by the number of nodes in the networks.  We compare the performance of our MCMC sampler (purple) and the highest scoring (MAP) DAG (red) from the final search space to the PC algorithm (blue) and GES (green), along with the MCMC sampler from a search space of the PC algorithm skeleton expanded to include one additional parent (pink).}
  \label{fig:SHD10n}
\end{figure}

\subsection{Direction inference}

Along with inferring the undirected skeleton, we also assess the performance in recovering the correct directions and compute the structural Hamming distance (SHD) between the true generating DAG and those inferred by the different methods.  The SHD is the minimum number of edge additions, deletions or reversals to transform one graph into another. In all cases we convert to CPDAGs before computing the distances. 

To visualise the SHD over different network sizes, we rescale by dividing by $n$: $\mathrm{SHDn}= \frac{\mathrm{SHD}}{n}$. The parameter values for each method were chosen to provide the smallest total median SHDn over the 8 simulation settings. For GES we used the penalisation $\lambda = 2\log(N)$ while for the PC algorithm we used a significance level of $\alpha=0.05$.  To condense the sample of DAGs from our MCMC schemes to a single graph, we converted the sample to CPDAGs and retained edges occurring with a posterior probability greater than $0.6$.  The result for targeting a MAP DAG correspond to the highest scoring DAG discovered in the final search space, again transformed into a CPDAG.

\begin{figure}[t]
  \centering
  \includegraphics[width=0.75\textwidth]{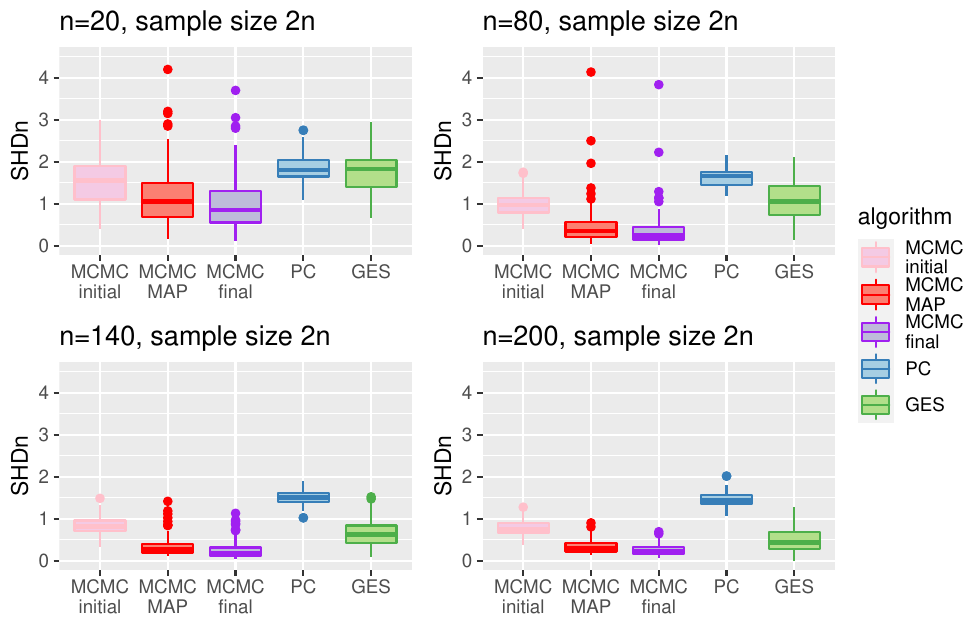}
  \caption{The performance in recovering the CPDAG.  The graph is as \fref{fig:SHD10n} but with a smaller sample size of $N=2n$ showing the SHD divided by the number of nodes for GES (green), the PC algorithm (blue), the initial and final MCMC samplers (pink and purple) and from the highest scoring DAG found using the MCMC schemes (red).}
  \label{fig:SHD2n}
\end{figure}

The results (\frefs{fig:SHD10n} and \ref{fig:SHD2n}) again show a strong improvement of our MCMC approach over the alternative algorithms.  Sampling and performing Bayesian model averaging also offers a consistent advantage over taking a MAP point estimate. 

\subsection{Other DAG types}

\begin{figure}[t]
  \centering
  \includegraphics[width=0.75\textwidth]{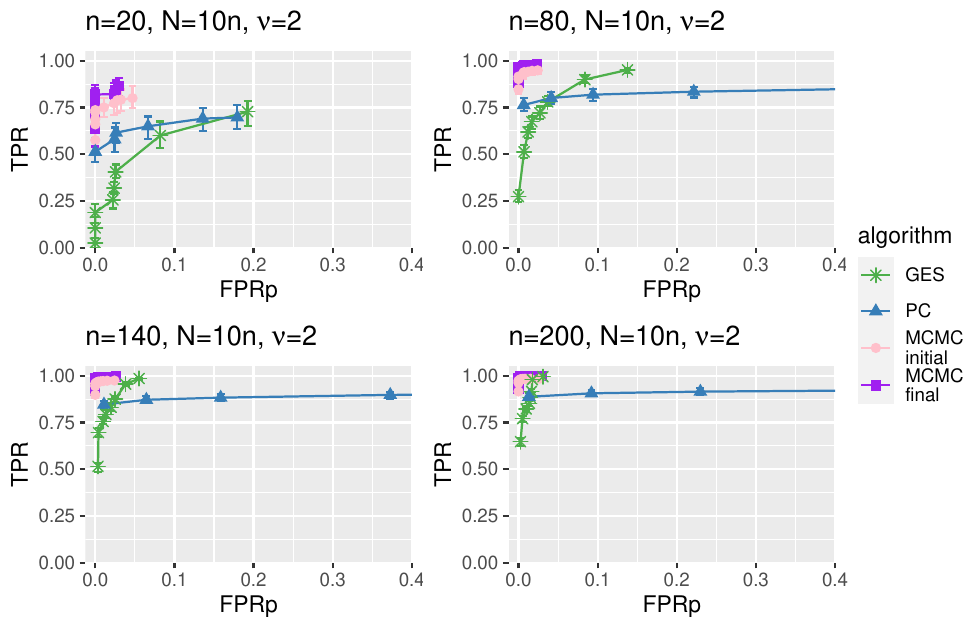}
  \caption{The performance in recovering the underlying DAG skeleton for Erd\H{o}s-R\'enyi DAGs with 2 parents on average.  We compare the performance of our MCMC scheme on its initial (pink circles) and final (purple squares) search spaces, to the PC algorithm (blue triangles) and GES (green stars).}
  \label{fig:ROCer2}
\end{figure}

For graphs again of size $n\in\{20, 80, 140, 200\}$ with a sample size of $N=10n$ and an average parent set size of 2, which we denote by $\nu=2$, we weaken the edge strengths and sample them uniformly in the range $[0.1, 1]$. In comparison to previous simulations (\fref{fig:ROC10n}) we observe (\fref{fig:ROCer2}) an overall decrease in FPs, with a worsening of GES relative to the improvement in the TPR of the PC algorithm). With a better starting search space, our initial MCMC sample already performs very well, with a consequently small improvement from the subsequent iterative search and sampling. For the smallest network size, some edges are however missed out, but overall  there is again a strong advantage for our MCMC approach.

\begin{figure}[t]
  \centering
  \includegraphics[width=0.75\textwidth]{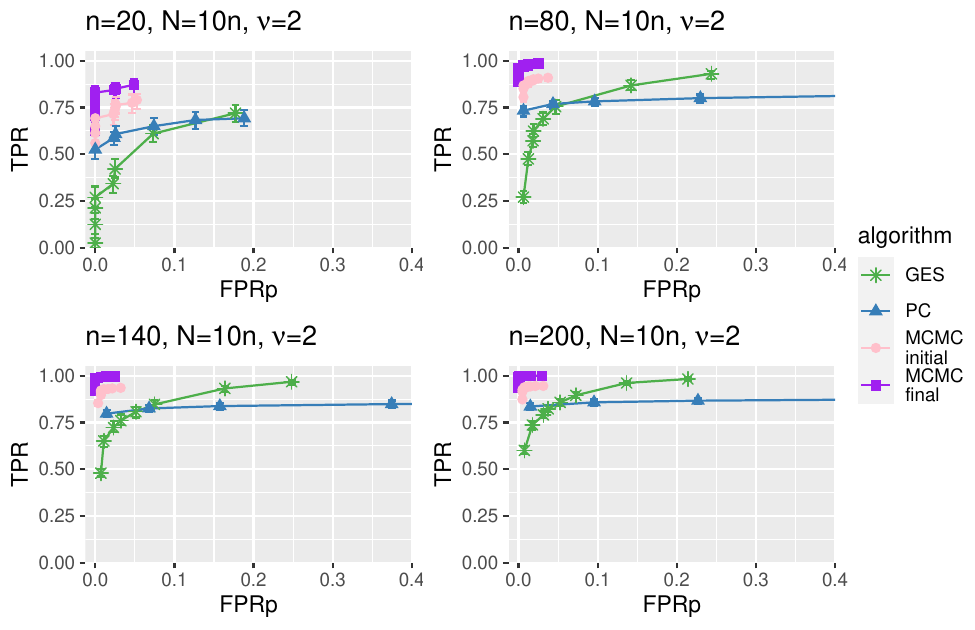}
  \caption{The performance in recovering the underlying DAG skeleton for Barab\'asi-Albert DAGs with 2 parents on average.}
  \label{fig:ROCbarab2}
\end{figure}

\begin{figure}[h!]
  \centering
  \includegraphics[width=0.75\textwidth]{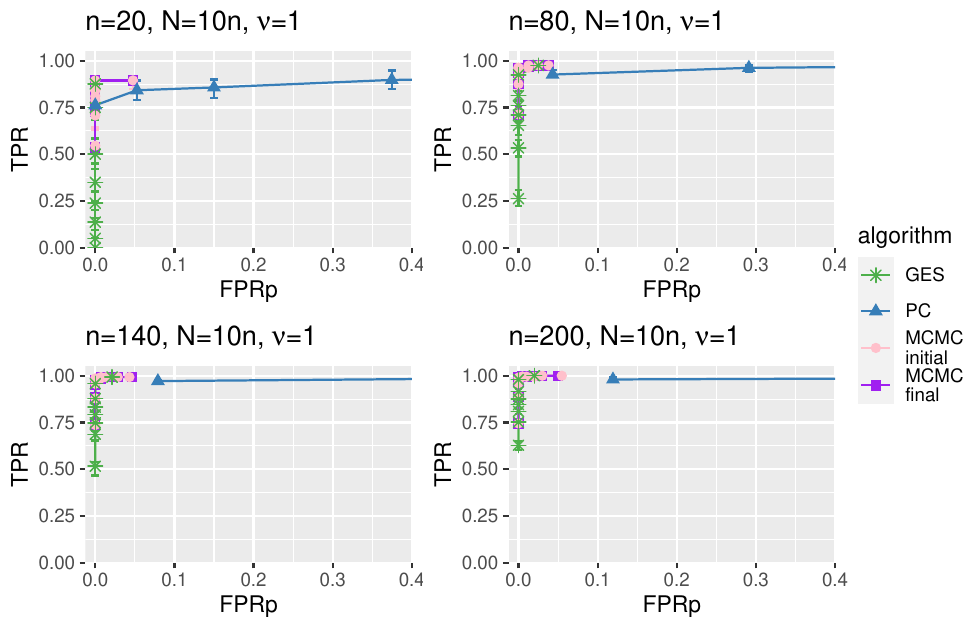}
  \caption{The performance in recovering the underlying DAG skeleton for Barab\'asi-Albert DAGs with 1 parent on average.}
  \label{fig:ROCbarab1}
\end{figure}

We next sample Barab\'asi-Albert scale-free or power-law DAGs, again with an average of 2 parents and a preferential attachment parameter of 0.4. We halve the default PC algorithm significance level to $\min(0.2, \frac{10}{n})$ to avoid it returning too many FPs. The results (\fref{fig:ROCbarab2}) are quite similar, although with a slight decrease in the performance of the PC algorithm and a notable worsening for GES for the larger networks. By sparsifying the graphs and setting the average number of parents to 1 instead, our MCMC scheme and GES perform very well, while the PC algorithm captures most true edges at the cost of lots of FPs (\fref{fig:ROCbarab1}).


\begin{figure}[t]
  \centering
  \includegraphics[width=0.75\textwidth]{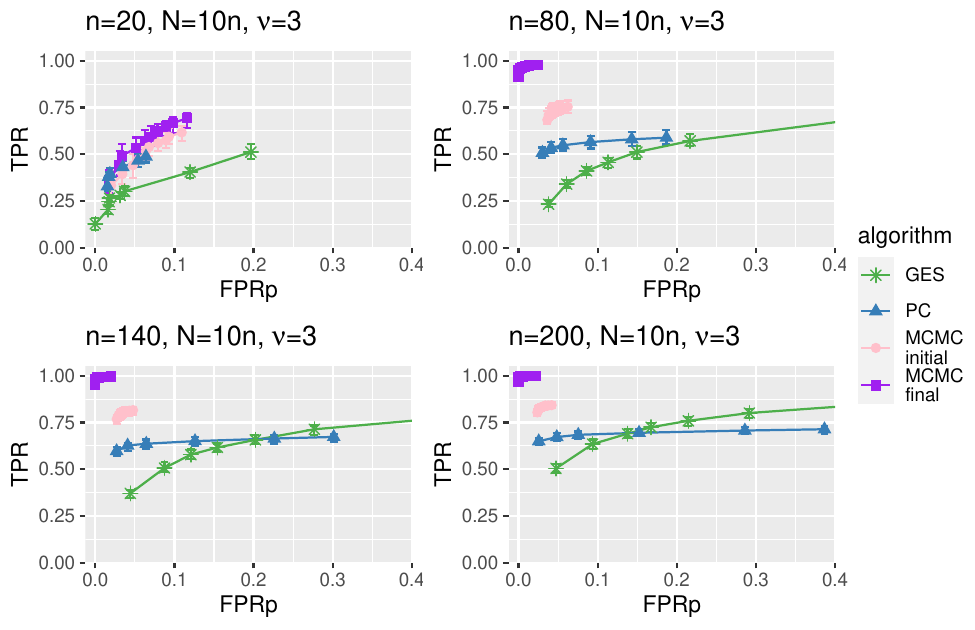}
  \caption{The performance in recovering the underlying DAG skeleton for a pair of Erd\H{o}s-R\'enyi islands with an interconnectivity coefficient of 0.1 and 3 parents on average.  The MCMC performance on a search space artificially including the true DAG is illustrated with grey diamonds.}
  \label{fig:ROCinterer3}
\end{figure}

Increasing the density instead to have an average of 3 parents per node, and having a network made up of two Erd\H{o}s-R\'enyi islands with an interconnectivity parameter of 0.1 (\fref{fig:ROCinterer3}), we observe for the larger networks that our MCMC scheme still manages to obtain nearly perfect recovery. In contrast, the PC algorithm struggles to find even half of them while GES finds some more, but at the cost of a large number of false positives.  Both alternatives perform substantially worse than MCMC.

\begin{figure}[h!]
  \centering
  \includegraphics[width=0.75\textwidth]{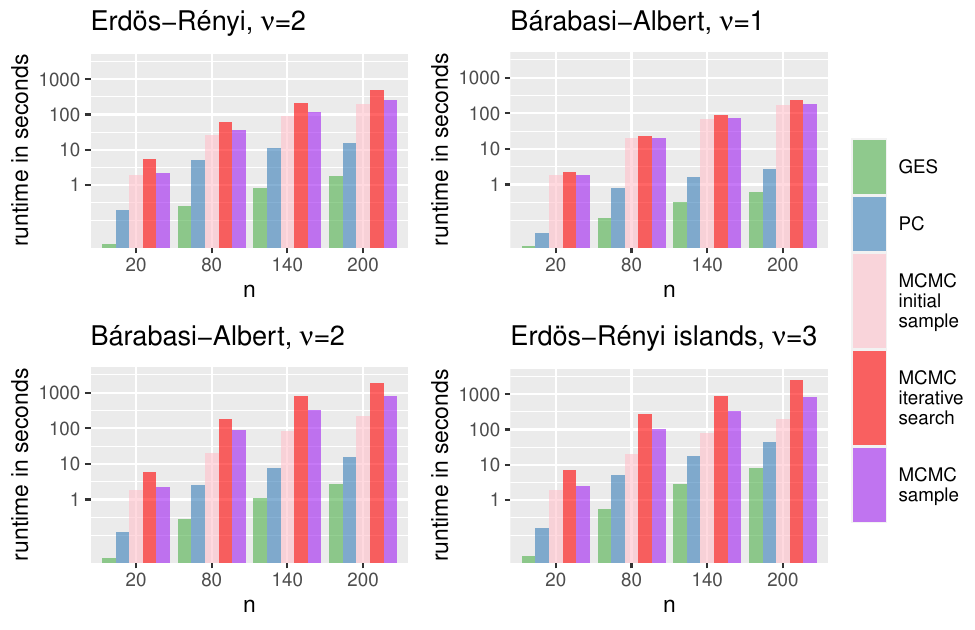}
  \caption{The average runtimes of finding the initial search space with the PC algorithm (blue), iteratively improving the search space (red) and the final MCMC sampling (purple) for the four different network settings.}
  \label{fig:runtime}
\end{figure}

The runtimes of GES are roughly an order of magnitude faster than the finding the initial search space with the PC algorithm, which is in turn roughly an order of magnitude faster than the MCMC sampling (\fref{fig:runtime}). Iteratively improving the search space is generally more costly than sampling, depending on the number of iterations needed. As the final search space can have a strong effect on the performance for denser networks (\fref{fig:ROCinterer3}), improvements in the speed of finding it could also be beneficial.  The actual runtimes of the full algorithm averages around 10 seconds for 20 nodes, around 10 minutes for 80 nodes, and up to a few hours for 200 nodes.

With their faster runtimes, the apparent scalability of the PC algorithm or GES to larger networks may appear tantalising. However, our study shows that their accuracy is very limited away from large sample size asymptotic regimes, so that the validity of the networks inferred remains dubious and caution in their interpretation is warranted.

\subsection{Categorical simulations} \label{catsim}

\begin{figure}[t]
  \centering
  \includegraphics[width=0.75\textwidth]{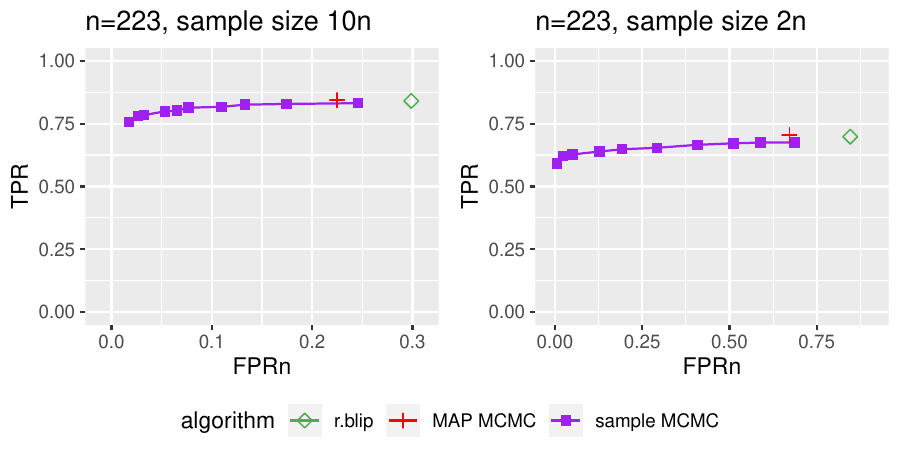}
  \caption{The accuracy in learning the ANDES network.  We compare the performance of \pkg{r.blip} to the MAP DAG search of \BiDAG and its posterior sampling.}
  \label{fig:andes}
\end{figure}

Comparing to the order-based search implemented in \pkg{r.blip} \citep{scanagatta2015learning}, requires categorical data.  First we considered the ANDES network with $n=223$ nodes from which we generated random samples \citep[with the \pkg{bnlearn} package;][]{bnlearn} for two sample sizes $N=2n$ and $N=10n$. With the intention to make a fair comparison we fixed the score parameters to those in \pkg{r.blip} and used the same runtime for both methods (1500 seconds) finding similar scores (slightly higher for \BiDAG when $N=2n$: -42844.62 against -42856.39; and slightly lower for $N=10n$: -209833.8 against -209811.9).  Irrespective of the score, we find a much better structure with \BiDAG with notably fewer false positive edges (\fref{fig:andes}).   Moreover, by accounting for the uncertainty in the edges and considering a posterior sample as opposed to a point estimate, we can drastically reduce the number of false positive edges while retaining true edges.  

\begin{figure}[t]
  \centering
  \includegraphics[width=0.5\textwidth]{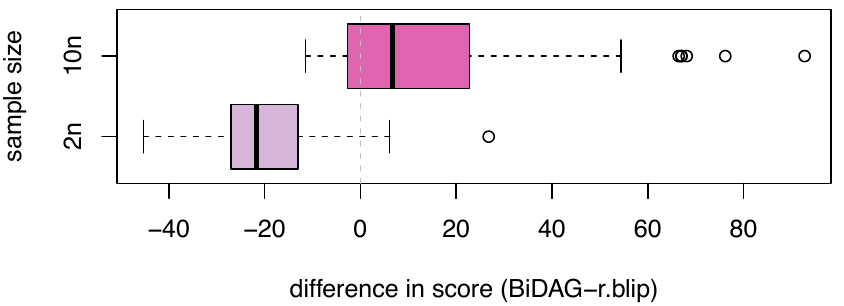}
  \caption{The relative log scores of the highest scoring DAG returned by \pkg{r.blip} and by \BiDAG for random networks of $n=100$ nodes.  The runtime for \pkg{r.blip} was fixed to 240 seconds, while \BiDAG had an average time of 150 seconds.}
  \label{fig:binsimscore}
\end{figure}

\begin{figure}[t]
  \centering
  \includegraphics[width=0.75\textwidth]{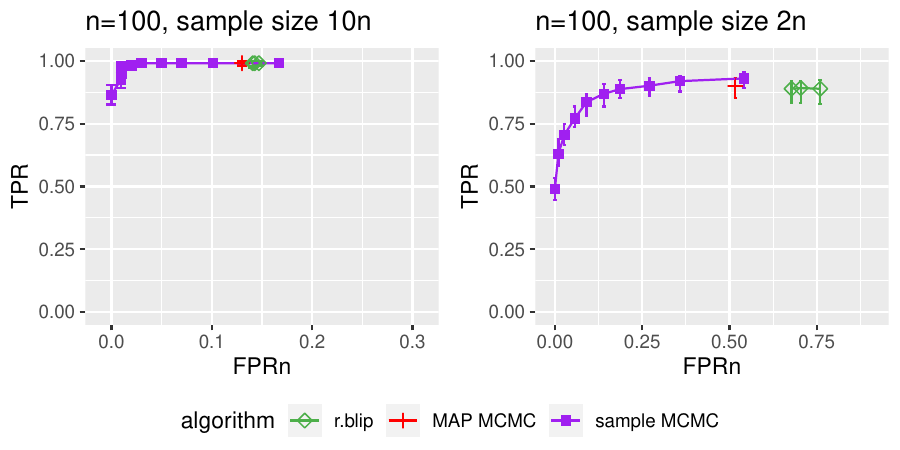}
  \caption{The accuracy in learning random networks with $n=100$ nodes.  We compare the performance of \pkg{r.blip} to the MAP DAG search of \BiDAG and its posterior sampling. The 3 points (left to right, with increasing values of FPR, but no appreciable change in TPR) for \pkg{r.blip} corresponds to runtimes of 60, 120 and 240 seconds, while \BiDAG had an average time of 150 seconds for MAP discovery, and 30 seconds for sampling.}
  \label{fig:binsim}
\end{figure}

A similar pattern is repeated in a larger scale simulation of 100 binary power-law networks (with 1 parent on average and a maximum of 5). In terms of score, we routinely find higher scoring graphs at the larger sample size, but mostly lower scores at the lower sample size (\fref{fig:binsimscore}). Considering the accuracy however highlights that running \pkg{r.blip} for longer times solely increases the false positive rate with no improvement in finding true edges (\fref{fig:binsim}). Therefore we may conclude that the higher score at the lower sample size is driven by false positives and does not lead to a more accurate graph structure. Furthermore, the real advantage of \BiDAG comes again from the ability to sample, and as a byproduct using a posterior threshold to remove false positive edges, which heavily improves the accuracy in network inference.


\section{Categorical data} \label{catdata}

For categorical data, we employ the BDe score \citep{hg95}.  For ease of presentation, we detail the binary case here.  For any node $X_i$, its contribution to the score involves computing the number of times $X_i$ takes the value 1, or the value 0, for each of the $2^K$ possible configurations of its $K$ parents $\Pa_i$.   All the parents for each of the $\Nobs$ observations must be run through in a complexity of $O(K\Nobs)$.  As there is a parameter associated with each of the $2^K$ parent configurations, we assume $\Nobs \gg 2^K$.  Building the score table of node $X_i$ by naively running through all parent configurations would take $O(K2^K\Nobs) \gg O(K4^K)$.

However the parent configurations are connected via the poset structure of \fref{possepic}, so we can build the score table more efficiently by only looking at the raw data once.  For the BDe score, for the case when all $K$ parents are present we build the vectors $\Nbold_{1}^{i}(\searchdgpa^{i})$ and $\Nbold_{0}^{i}(\searchdgpa^{i})$ whose $2^K$ elements count the number of times $X_i$ takes the value 1 and 0 for each parent state, in time $O(K\Nobs)$.  We employ a binary mapping of the parent states to elements of the vectors using
\begin{equation}
\sum_{j=1}^{\vert \setVarexample \vert }I(Z_j =1)2^{j-1}
\end{equation}
where for the full set of parents, $\setVarexample=\searchdgpa^{i}$.  When we remove one of the parents to compute the score table entry at layer $(K-1)$ in the poset of \fref{possepic} we simply combine elements of the vector $\Nbold_{\{1,0\}}^{i}$ where the removed parent takes the value 0 with the corresponding elements where it takes the value 1.  In general we can create the vectors at each level from any connected at a higher level with
\begin{equation} \label{eqcombinebinaryvectors}
\Nbold_{\{1,0\}}^{i}(\setVarexample \setminus \setVarexample_j)[t] = \Nbold_{\{1,0\}}^{i}(\setVarexample)[v(t,j)] + \Nbold_{\{1,0\}}^{i}(\setVarexample)[v(t,j)+2^{j-1}]
\end{equation}
where the square brackets indicate the elements of the vectors and we employ the mapping
\begin{equation}
v(t,j) = t + \left(\left\lceil \frac{t}{2^{j-1}} \right\rceil -1 \right)2^{j-1}
\end{equation}
From the pair of vectors for any set of permissible parent nodes $\Nbold_{\{1,0\}}^{i}(\setVarexample)$ we can compute the entry in the score table according to the BDe score \citep{hg95}
\begin{equation} \label{BDescoreeqn}
\score^{i}_{f(\setVarexample)} = \sum_{t=1}^{2^{m}} \frac{\Gamma\left(\frac{\chi}{2^{m}}\right)}{\Gamma\left(\frac{\chi}{2^{m+1}}\right)\Gamma\left(\frac{\chi}{2^{m+1}}\right)} \frac{\Gamma\left(\Nbold_{1}^{i}(\setVarexample)[t]+\frac{\chi}{2^{m+1}}\right)\Gamma\left(\Nbold_{0}^{i}(\setVarexample)[t]+\frac{\chi}{2^{m+1}}\right)}{\Gamma\left(\Nbold_{1}^{i}(\setVarexample)[t]+\Nbold_{0}^{i}(\setVarexample)[t]+\frac{\chi}{2^{m}}\right)}
\end{equation}
with $m=\vert \setVarexample \vert$ and $\chi$ the hyperparameters of the beta distributions which correspond to pseudocounts in the score.

\begin{algorithm}[t]
\caption{Obtain the scores of all parent sets for binary data}\label{binarydataalg}
\begin{algorithmic}
\State \textbf{Input} The power set network of the $K$ permissible parents of variable $i$
\State \textbf{Input} The count vectors of the full parent set $\Nbold_{\{1,0\}}^{i}(\searchdgpa^{i})$
\State Label the network nodes $Y_{f(\setVarexample)}$ for each $\setVarexample$ in the power set
\State Compute the score $\score^{i}_{2^{K}-1}$ at layer $K$ from the $\Nbold_{\{1,0\}}^{i}(\searchdgpa^{i})$ \Comment{\eref{BDescoreeqn}}
\For{$l=K-1$ to $0$} \Comment{layer number} 
\ForAll{$\{j \in \mbox{ layer } l\}$}
\State Choose any child in the network
\State Compute $\Nbold_{\{1,0\}}^{i}(f^{-1}(j))$ from the child \Comment{\eref{eqcombinebinaryvectors}}
\State Compute $\score^{i}_{j}$ from the $\Nbold_{\{1,0\}}^{i}(f^{-1}(j))$ \Comment{\eref{BDescoreeqn}}
\EndFor
\EndFor
\State \textbf{return} Table of scores: $\score^{i}_j$, $j=0,\ldots,(2^{K}-1)$
\end{algorithmic}
\end{algorithm}

Repeating the creation of the count vectors and computation of the score by moving up the layers in the poset of \fref{possepic}, as summarised in \alref{binarydataalg}, we efficiently build the score table for each node in the data.  For each term at layer $l$ we look at vectors from the layer above of size $2^{l+1}$ so that filling out the score tables takes $O(3^K)$.  Combining with the initial step leads to an overall complexity of $O(\max\{K\Nobs,3^{K}\})$ which is a significant improvement on the naive implementation of $O(\max\{K2^{K}\Nobs,K4^{K}\})$.

For categorical data, the same approach is followed, although with mixed radix indexing for different sized categories rather than the simple binary mapping discussed above.  With more possible states, the complexity also increases.  For example if all categories have $C$ levels, the complexity is $O((C+1)^K)$.

\clearpage

\vskip 0.2in
\bibliography{mcmcdags}

\end{document}